\documentclass[lettersize,journal]{IEEEtran}
\usepackage{amsmath,amsfonts}
\usepackage{algorithmic}
\usepackage{algorithm}
\usepackage{array}
\usepackage[caption=false,font=normalsize,labelfont=sf,textfont=sf]{subfig}
\usepackage{textcomp}
\usepackage{stfloats}
\usepackage{url}
\usepackage{verbatim}
\usepackage{graphicx}
\usepackage{cite}
\usepackage{booktabs}
\usepackage{multirow}
\usepackage{makecell}
\usepackage{xcolor}
\usepackage{caption}
\usepackage{threeparttable}
\usepackage[colorlinks = true,
            linkcolor = blue,
            urlcolor =blue,
            citecolor = blue,
            anchorcolor = blue]{hyperref}

\begin{document}
\title{Detecting Children with Autism Spectrum Disorder \\ based on Script-Centric Behavior Understanding \\ with Emotional Enhancement}

\author{Wenxing Liu,Yueran Pan,Dong Zhang, Hongzhu Deng, Xiaobing Zou and  Ming Li, senior member
\thanks{ This research is funded in part by the Guangdong Science and Technology Plan (2023A1111120012),  and National Natural Science Foundation of China (62171207, 62173353). (Corresponding author: Ming Li.)}
\thanks{The collection and analysis of the clinical database was approved by the Third Affiliated Hospital of Sun Yat-sen University Institutional Review Board (IRB No. [2018]02-196-01 and [2020]02-118-03) and Duke Kunshan University Institutional Review Board (IRB No. 2022ML065,2023LM159, 2025LM039).  Access to the audio-visual data remains restricted, but we are prepared to share and contribute towards collaborative efforts concerning the available behavior scripts data.} 
\thanks{Wenxing Liu , Yueran Pan and Ming Li are with the School of Computer Science, Wuhan University, Wuhan, China, 430072, and the Digital Innovation Research Center, Duke Kunshan University, Kunshan, China, 215316, (e-mail: \href{mailto:wenxing.liu@dukekunshan.edu.cn}{wenxing.liu@dukekunshan.edu.cn}; \href{mailto:panyr.math@whu.edu.cn}{panyr.math@whu.edu.cn}; \href{mailto:ming.li369@dukekunshan.edu.cn}{ming.li369@dukekunshan.edu.cn}). Dong Zhang is with the School of Electronics and Information Technology, Sun Yat-sen University, Guangzhou 510006, China (e-mail: \href{mailto:zhangd@mail.sysu.edu.cn}{zhangd@mail.sysu.edu.cn}). Hongzhu Deng and Xiaobing Zou are with the Child Development and Behavior Center, Third Affiliated Hospital of Sun Yat-sen University, No.600 Tianhe Road, Guangzhou, China, 510630, (email: \href{mailto:denghongzhu@foxmail.com}{denghongzhu@foxmail.com}; \href{mailto:zouxb@mail.sysu.edu.cn}{zouxb@mail.sysu.edu.cn})}}

\markboth{IEEE TRANSACTIONS ON AFFECTIVE COMPUTING}%
{Shell \MakeLowercase{\textit{et al.}}: A Sample Article Using IEEEtran.cls for IEEE Journals}

\maketitle

\begin{abstract}
The early diagnosis of autism spectrum disorder (ASD) is critically dependent on systematic observation and analysis of children's social behaviors. While current methodologies predominantly utilize supervised learning approaches, their clinical adoption faces two principal limitations: insufficient ASD diagnostic samples and inadequate interpretability of the detection outcomes. This paper presents a novel zero-shot ASD detection framework based on script-centric behavioral understanding with emotional enhancement, which is designed to overcome the aforementioned clinical constraints. The proposed pipeline automatically converts audio-visual data into structured behavioral text scripts through computer vision techniques, subsequently capitalizing on the generalization capabilities of large language models (LLMs) for zero-shot/few-shot ASD detection. Three core technical contributions are introduced: (1) A multimodal script transcription module transforming behavioral cues into structured textual representations. (2) An emotion textualization module encoding emotional dynamics as the contextual features to augment behavioral understanding. (3) A domain-specific prompt engineering strategy  enables the injection of clinical knowledge into LLMs. Our method achieves an F1-score of 95.24\% in diagnosing ASD in children with an average age of two years while generating interpretable detection rationales. This work opens up new avenues for leveraging the power of LLMs in analyzing and understanding ASD-related human behavior, thereby enhancing the accuracy of assisted autism diagnosis.
\end{abstract}


\section{Introduction}
\label{sec:intro}
\IEEEPARstart{A}{utism} is a neurodevelopmental disorder characterized by impairment in communication and social interaction, restricted interests and stereotyped behaviors \cite{lord2018autism,regier2013dsm}. It has 

\begin{figure}[ht]
\centering
\centerline{\includegraphics[width=8.5cm]{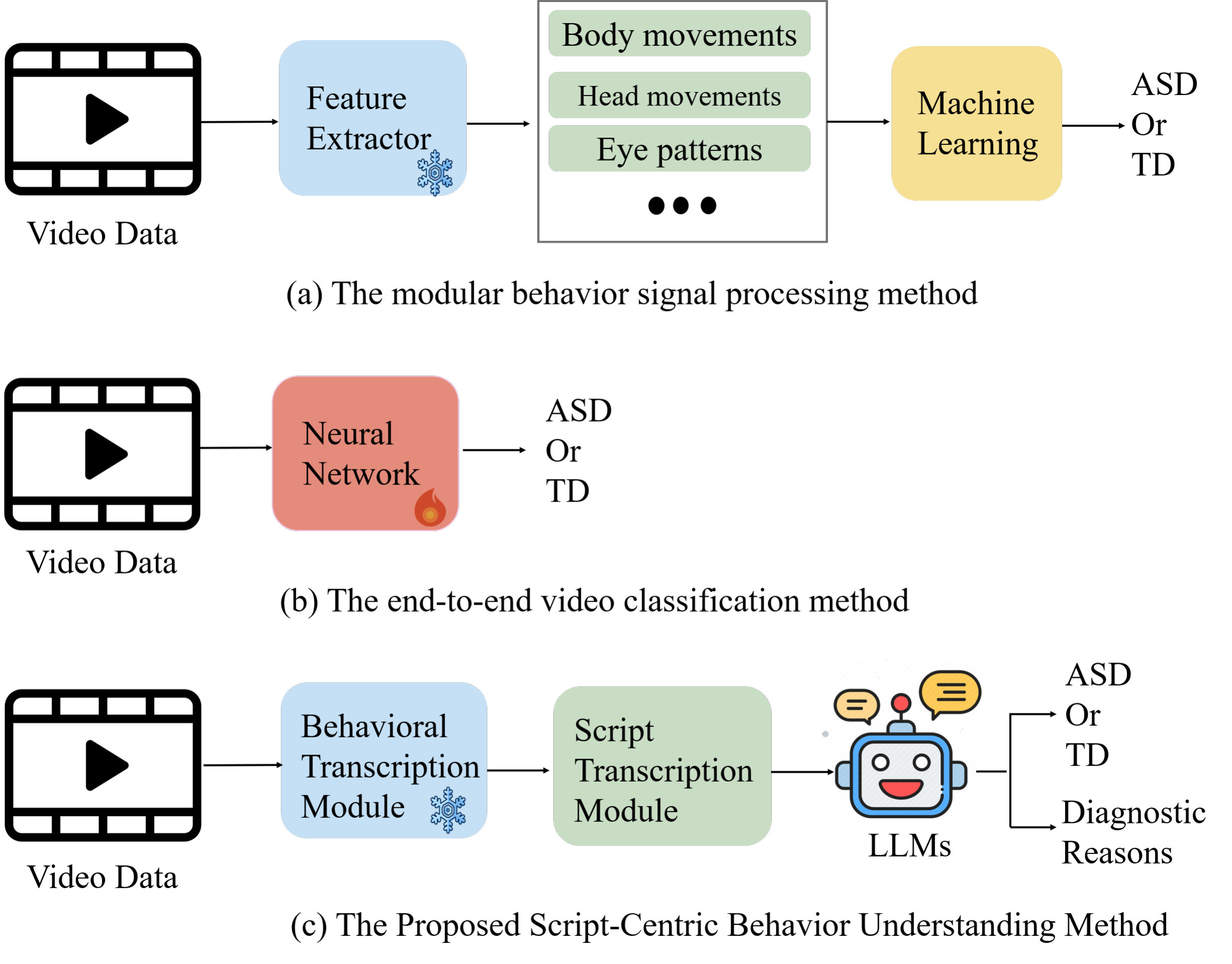}}
\caption{Pipeline comparing (a) behavioral signal processing method , (b) raw-video-based method and (c) The proposed script-centric behavior understanding (SCBU) method}
\label{fig0}
\end{figure}

\noindent brought substantial psychological and financial burdens on affected families while leading to broader societal impacts \cite{lord2022lancet}.

Numerous studies show that behavioral interventions for individuals with ASD can effectively mitigate core symptoms when initiated during key developmental periods \cite{reichow2012overview}. Accurate early diagnose is essential for the subsequent intervention. However, the conventional gold-standard diagnostic tools, like the Autism Diagnostic Interview-Revised (ADI-R) \cite{Lord1994} and the Autism Diagnostic Observation Schedule (ADOS) \cite{Lord1989}, typically rely on interviews or behavioral observations, which require highly experienced clinicians and is quite time-consuming. These methodologies exhibit inherent subjectivity due to variabilities in interpretation and differences in individual expertise. Hence, the implementation of these diagnostic protocols continues to pose significant challenges, particularly  in resource-constrained healthcare regions. This underscores the significant and critical need for developing an objective, accurate and interpretable early screening tool for ASD \cite{uddin2024deep}.

To address these problems, existing research have demonstrated the efficacy of artificial intelligence techniques in ASD detection using multimodal sensor data, e.g. functional Magnetic Resonance Imaging (fMRI) data \cite{Huang2021,Park2023,Liu2024}, Electroencephalogram data (EEG) data \cite{Ardakani2022,neo2023resting,wilkinson2024developmental}, behavioral observation audio-visual data \cite{Zhang2023,cheng2023computer,Deng2024}, etc. This paper focuses on automated ASD detection through audio-visual data analytics. Conventional behavior data based methods can be divided into two main categories, as shown in Fig.~\ref{fig0} (a) and (b). The first is based on modular behavioral signal processing. These methods extract the child's behavioral features (e.g., gaze patterns \cite{zhou2024gaze}, head movements \cite{li2022automatic}, body motions \cite{kojovic2021using}, audio prosody and text transcripts \cite{Fan2024}, facial expression and hand gestures \cite{Li2019}, etc.) and then analyze these features by backend machine learning algorithms. The second is based on end-to-end raw video classification, which directly learns the mapping relationship between video and labels through deep learning models \cite{prakash2023computer, rani2024activity, xia2024identifying}. Both methods provide quantitative and objective diagnostic tools for ASD detection. However, these supervised learning methods need to be trained by a large amount of data, especially for the end-to-end video classification approach. Hence,  the scarcity of ASD data limits their accuracy in practice. Moreover, most of the aforementioned methods are limited to making binary predictions and lack detailed explanations supporting those detection outcomes.

In recent years, researchers have gradually utilized LLMs \cite{Wu2023can, zhang2024generalist, guevara2024large} to explore new approaches for medical diagnosis. Compared to traditional AI algorithms, LLMs exhibit two primary advantages for clinical applications \cite{Zhou2023,Thirunavukarasu2023,Omiye2024}. First, their ability to leverage extensive medical knowledge enables reasoning and contextual comprehension, forming an interpretable foundation for diagnostic decision-making. This capability establishes a crucial foundation for ensuring the interpretability of diagnostic decisions. Second, their inherent prior knowledge supports zero-shot and few-shot learning, reducing dependence on clinical training data.

While LLMs show potential for medical diagnostics, three fundamental challenges persist in processing ASD-specific audio-visual behavioral signals: 1) Modality adaptation gap. Current LLMs excel at text and image processing \cite{gpt4o, Anthropic, dubey2024llama, yang2024qwen2, Kimi, guo2025deepseek}, but face inherent limitations in accurate and robust cross-modal alignment between audio-video inputs and textual representations \cite{li2023blip,zhang2023video}, creating fundamental barriers to biomedical behavioral signal processing. 2) Domain prompt Design. While prompt engineering significantly enhances LLM capabilities for domain-specific tasks \cite{white2023prompt}, the systematic design of effective prompts and the optimal infusion of ASD-related domain knowledge still remain important research questions. 3) Emotional dynamic Modeling. Emotional dynamic is a critical biomarker for ASD detection \cite{begeer2008emotional, mazefsky2013role, rashidan2021technology}, a key challenge lies in modeling emotional dynamic as a textual feature that enhances LLMs-based ASD detection.

Therefore, this paper introduces a novel pipeline for ASD detection using LLMs, which can determine ASD and provide explanations. First, to solve the problem of mismatch between audio-visual behavior data and the input modality of LLMs, we convert the video content (e.g. the characters' gestures, head poses, body' movements, facial expressions, speech content, gaze, and etc.) as time-stamped text inspired by movie scripts. Specifically, we use a behavioral transcription module to convert the video content into human behavioral logs, and a script transcription module is designed to process these behavioral logs into natural language texts. Second, to improve the accuracy of LLM for ASD detection, we created a domain prompt module to incorporate ASD domain knowledge. Finally, we design the emotion textualization module to enhance LLMs' understanding of the emotional dynamics in ASD  detection. Our main contributions can be summarized as follows:

\begin{itemize}
    \item[$\bullet$] To the best of our knowledge, our approach is the first to introduce LLMs for detecting ASD from audio-visual data, laying the foundation for the exploration of LLMs in this domain. 
    \item[$\bullet$] In order to accurately describe the behavioral data and utilize the domain knowledge, we propose a script transcription module and a domain prompt module. They build a bridge between audio-visual data and LLMs, facilitating the development of multimodal ASD detection.
    \item[$\bullet$] We design an emotion textualization module to add emotional dynamics that are usually ignored in the behavioral script, further enhancing the detection accuracy.
\end{itemize}

\section{Related Work}
\subsection{Modular Behavior Signal Processing Methods for ASD Detection}
To overcome ASD data scarcity, some approaches leverage cross-domain behavioral signal processing modules for feature extraction, followed by ASD-targeted modeling with domain knowledge. Hashemi et al. \cite{hashemi2018computer} developed a mobile screening system that employs cinematic stimuli to elicit quantifiable social responses  (e.g., name recognition, joint attention, affective reciprocity) for automated ASD classification via behavioral pattern analysis. Negin et al. \cite{negin2021vision} implemented a Bag-of-Visual-Words (BoVW) framework to extract local descriptors from video data, employing multiple machine learning classifiers for automated ASD screening. Zhang et al. \cite{zhang2023discriminative} implemented a 3D spatiotemporal facial analysis pipeline and a few-shot learning strategy to evaluate discriminative facial dynamics for ASD classification. Atyabi et al. \cite{atyabi2023stratification} developed a multimodal integration framework combining behavioral biomarkers—including eye movement scan paths, temporal information and pupil velocity—to differentiate ASD and Typically Developed (TD). Cheng et al. \cite{cheng2023computer} developed a computer-aided ASD detection system employing multimodal behavioral signal analysis. The system's multimodal behavioral transcription module and response parser recognizes audio-visual signals to identify child's behaviors, and a back-end machine learning model is trained based on the paradigm scores and behavioral features to provide assisted ASD detection. Nie et al. \cite{chen2024deep} formalized child-caregiver interactions through a Computational Interpersonal Communication Model (CICM) grounded in Theory of Mind (ToM), employing Markov decision processes to decode multimodal behavioral signals for early screening. While demonstrating diagnostic potential, these approaches predominantly require substantial domain-specific expertise and customized  behavioral signal processing module designs, exhibiting limited generalizability across diverse clinical scenarios.

\subsection{End-to-end video classification for ASD Detection}
Many researchers employ deep learning algorithms to directly model raw videos in an end-to-end manner. Li et al. \cite{li2020classifying} pioneered an LSTM-based deep learning architecture for automated ASD detection through raw video, specifically targeting discriminative gaze pattern classification between ASD and TD. Pandian et al. \cite{pandian2022detecting} developed the RGBPose-SlowFast network to automatically detect stereotypical motor behaviors in children with ASD, demonstrating the viability of multi-stream neural networks for automated ASD screening. Wei et al. \cite{wei2023vision} explored a hybrid architecture integrating handcrafted feature extractors with 3D convolutional neural networks for automated detection of stereotyped motor mannerisms in ASD. Chen et al. \cite{xia2024identifying} employ Longformer to establish the correlation of facial features in videos over time, aiming to learn a deep representation from dynamic facial data for ASD detection.  Asha et al. \cite{rani2024activity} developed a supervised contrastive learning framework to extract cross-dataset discriminative feature representations for ASD and TD, ultimately deploying an automated diagnostic classification pipeline leveraging raw video. However, current limitations in ASD audio-visual datasets critically constrain raw video classification techniques that demand extensive training data for reliable behavioral pattern recognition.

\begin{figure*}[htb]
    \centering
    \includegraphics[width=\textwidth,scale=1.00]{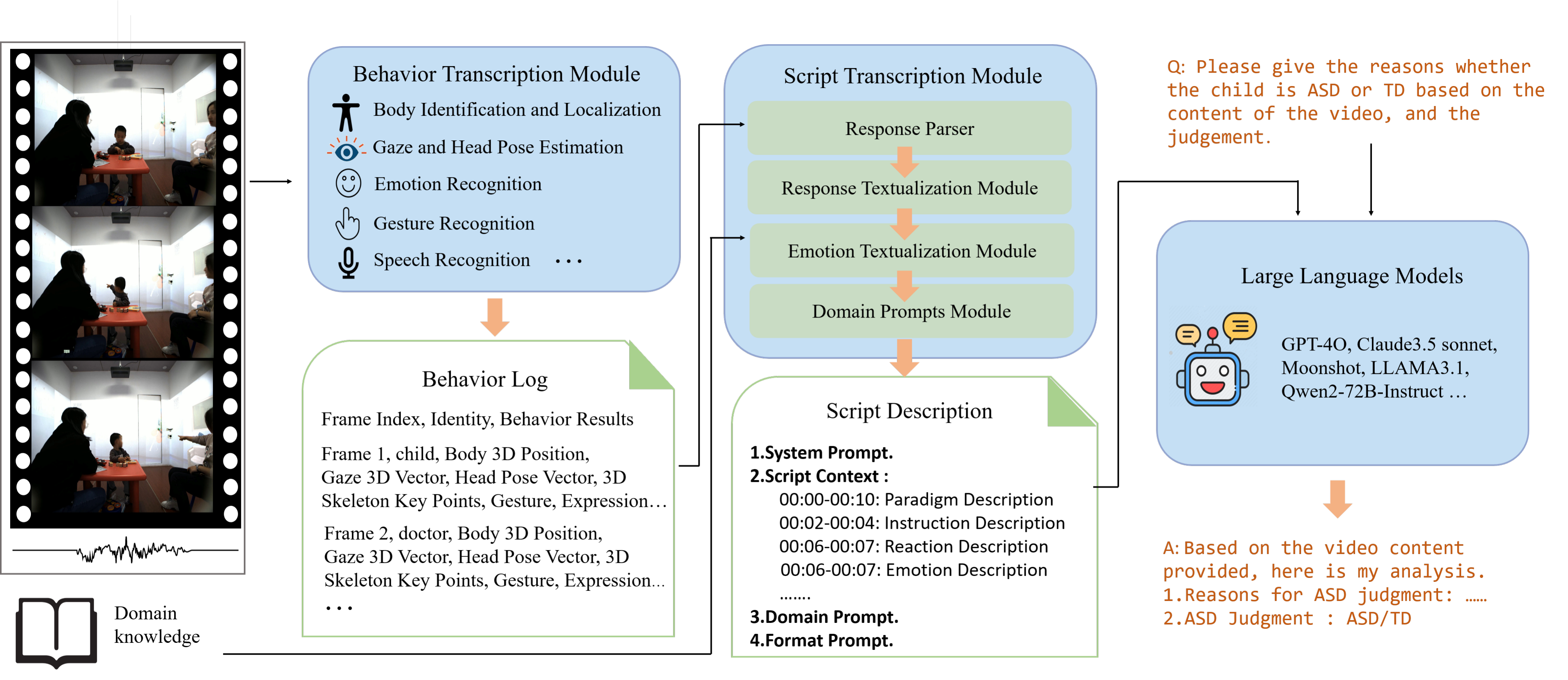}
    \caption{The overview of our proposed Script-Centeric Behavior Understanding(SCBD) framework. \textbf{Behavior Transcription Module} converts audio-video data into behavioral logs using multiple well-trained behavior signal processing models. \textbf{Scipt Transcription Module} textualizes Behavior Logs in steam and integrate domain prompt. \textbf{Large Language Models} are used to understand and anwser script content. } 
    \label{fig1}
\end{figure*}

\subsection{LLMs Based Methods For Medical Detection}
In recent years, LLMs have demonstrated advanced natural language processing capabilities, showing particular promise in clinical diagnostics for developing interpretable decision-making systems that balance accuracy and transparency. The Med-PaLM \cite{singhal2023large} is a medical LLM developed by the Google team. Its strong performance in answering medical questions demonstrates the extensive medical knowledge embedded in LLMs. Huatuo GPT \cite{wang2023huatuo} is an LLM for medical consulting built on the open-source LLaMa-7B model \cite{zeng2022glm}. The LLM learns structured and unstructured medical knowledge from the Chinese Medical Knowledge Atlas and can diagnose over 3,000 diseases.  Through a prompt engineering approach, Medprompt \cite{nori2023can} demonstrates that general LLMs achieve competitive performance even without domain-specific pretraining. The LLaVA-ASD \cite{deng2024hear} multimodal LLM is designed to detect social and repetitive behaviors through audio-visual cues. While multimodal LLMs can process audio-visual inputs directly, insufficient detail in behavioral descriptions generated by general visual understanding models lead to reduced performance.

\subsection{Emotion Based Methods for ASD Detection}
The DSM-V \cite{regier2013dsm}, published by the American Psychiatric Association, is the most authoritative and widely used manual for diagnosing mental disorders, including ASD. Its criteria emphasize significant deficits in the emotional domain, such as lack of facial expression. These challenges stem from impaired emotional expression and regulation mechanisms in ASD \cite{begeer2008emotional}.
This has driven the development of computational approaches targeting emotion-related deficits in ASD through computer-assisted methodologies. Sarabadani et al. \cite{sarabadani2018physiological} proposed a method to recognize emotional states by physiological signals automatically. The study indicated that children with ASD exhibited different responses compared to children with TD when viewing images of the same emotional valence. Piana et al. \cite{piana2019effects} developed an automatic emotion recognition system to support children with ASD in learning emotion recognition and expression through whole-body movements. Prakash et al. \cite{prakash2023computer} develop a framework for extracting motor behaviors, emotional states, and facial expressions from child-caregiver interactive videos. This multimodal integration of behavioral and affective data enables robust diagnostic frameworks for ASD. Rashidan et al. \cite{rashidan2023stimuli} verified that appropriate video stimuli can elicit emotional responses in children with ASD and also demonstrated significant differences in emotion regulation between children with ASD and TD.

In summary, given the limited audio-visual data in the ASD domain, using behavior signal processing modules, including emotion recognition modules,  to generate high quality behavior scripts and feeding them to LLMs with ASD domain knowledge for zero-shot or few-shot learning is a very promising area to explore for ASD detection.

\begin{figure*}[htb]
    \centering
    \includegraphics[width=\textwidth,scale=1.00]{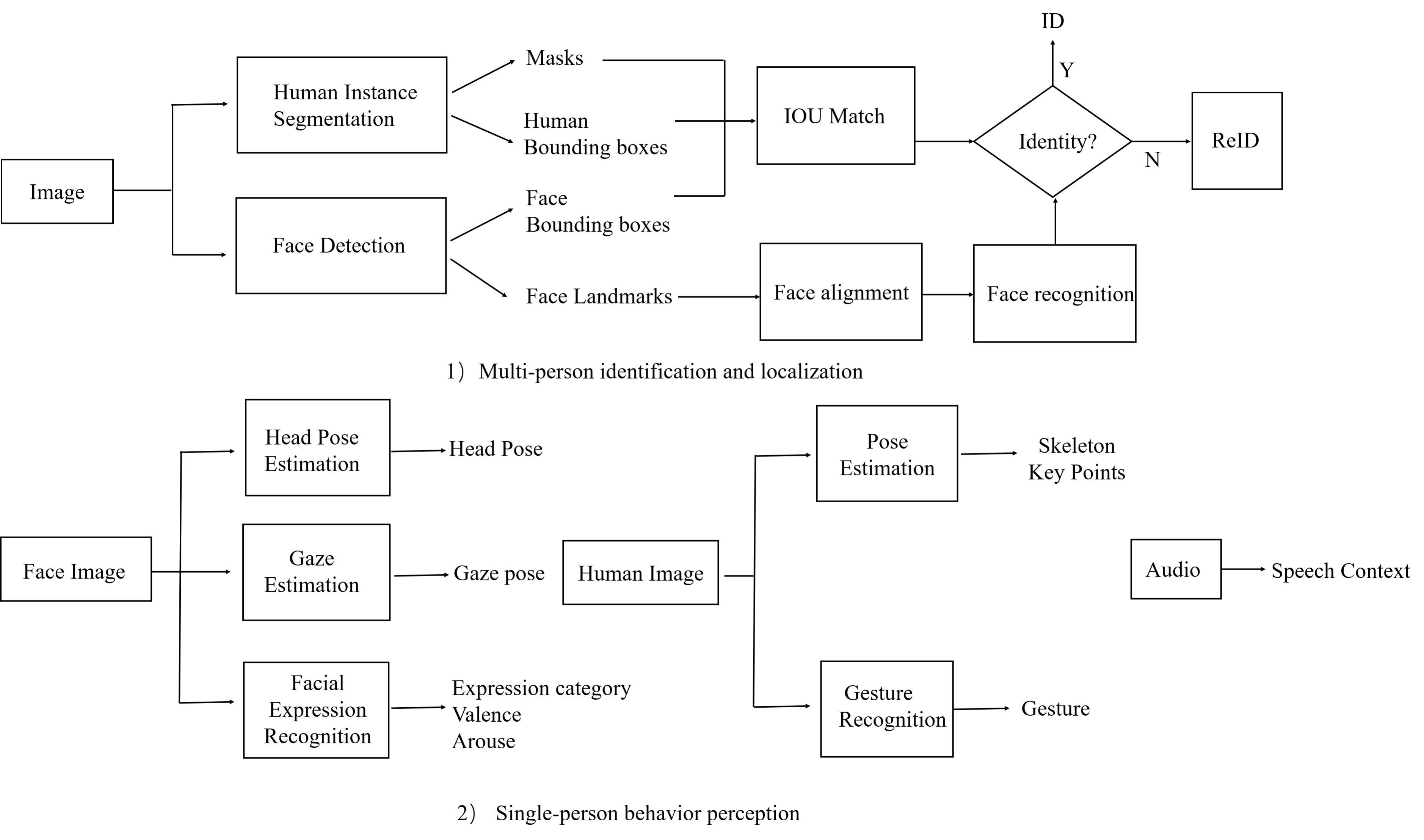}
    \caption{A two-stage pipeline in the behavioral transcription module. (1) \textbf{Multi-person identification and localization} is used to locate the location information and identify the participant in each frame. (2) \textbf{Single-person behavior perception} is used to perceive the behavioral information of each individual} 
    \label{fig2}
\end{figure*}

\section{Method}
Fig.~\ref{fig1} illustrates the overall framework of our proposed Script-Centric Behavior Understanding (SCBU) method, enabling LLMs to detect ASD from audio-visual behavior data automatically. The framework follows a sequential structure consisting of three modules: Behavioral Transcription Module, Script Transcription Module and LLMs. The Behavioral Transcription Module recognize basic human behaviors from audio-video data. The Script Transcription Module, newly proposed in this work, bridges the gap between behavioral logs and LLMs. It comprises four components: 1) The response parser module captures predefined response signs. 2) The response textualization module converts behavior activity into a script (text format). 3) The emotion textualization module enhances script details by adding appropriate emotional descriptions. 4) The domain prompts module combines script context, system prompt and ASD-related knowledge together to understand human behaviors better. Ultimately, we rely on pretrained LLMs to detect ASD and produce judgments by answering questions based on the script description.

\subsection{Behavior Transcription Module}
Following our previous work \cite{cheng2023computer}, the behavior transcription module can recognize the position, gesture, body movement, head pose, eye gaze, emotion, speech content of all individuals in each segment by combining audio and image models. As shown in Fig.~\ref{fig2}, the behavioral transcription module is divided into two phases to convert recorded audio-visual data into behavioral logs. 

The first stage, called the multi-person identification and localization stage, has the core objective of localizing and identitying different people in each frame. For each image frame, we first use an instance segmentation model (Solov2\cite{wang2020solov2}) to extract human body regions, including bounding boxes and masks. Moreover, it also goes through a face detection model (RetinaFace\cite{deng2020retinaface}) to localize face bounding boxes. To determine which body bounding box corresponds to the face bounding box, we use the smallest Intersection Over Union (IOU) for matching. Then, the face recognition model (ArcFace\cite{deng2019arcface}) IS used to distinguish the identity of characters. If face recognition is possible due to facial occlusion, we

\begin{table}[htp]
\setlength{\tabcolsep}{1.5mm}
\caption{Basic response events included in each assessment paradigm}
\vspace{0.5em}
\label{tab:paradigm}
\centering
\begin{tabular*}{\linewidth}{cccccc}
\toprule
\textbf{Paradigm} &\textbf{Look at Object}&\textbf{Point at Object} & \textbf{Smile}  & \textbf{Speak} & \textbf{Leave} \\ \midrule
RN & $\surd$ &         &         & $\surd$ &  \\
SS & $\surd$ &         & $\surd$ &         &         \\
IG & $\surd$ & $\surd$ &         &         &         \\
RJA& $\surd$ &         &         &         &         \\
IJA& $\surd$ & $\surd$ &         &         &         \\
SA & $\surd$ &         &         &         & $\surd$ \\ \bottomrule
\end{tabular*}
\end{table}

\begin{algorithm}[htb]
    \caption{Find Emotinonal Dynamic Points}
    \label{alg:findemo}
    \renewcommand{\algorithmicrequire}{\textbf{Input:}}
    \renewcommand{\algorithmicensure}{\textbf{output:}}
    \begin{algorithmic}[1]
        \REQUIRE The Valence sequence  $\left[a_1, a_2, \ldots, a_n\right]$ for children with ASD
        \ENSURE The emotional dynamic points $P_n$ , and its time interval $[t_{n_{-} s t a r t}, t_{n_{-} e n d}]$

        \STATE Calculate the first-order derivatives of the valence sequence , and obtain the first-order derivative sequence $\left[d_1, d_2, \ldots, d_n\right]$.

        \FOR{each $i \in [0,n]$}
            \IF { $d_n>0.2$ or $dn < 0.2$ }
                \STATE  Record the moment $t_n$ of the $P_n$, and the interval of dynamic $[t_n-0.5s,t_n+0.5s]$.
            \ENDIF
        \ENDFOR

        \WHILE{$P_n$ and $P_{n+1}$ have overlap}
            \STATE Merged into one continuous emotional segment: $P_n = [t_{n_{-} s t a r t}, t_{n+1_{-} e n d}]$.
            
        \ENDWHILE
        
    \end{algorithmic}
    
\end{algorithm}

\noindent use the person re-identification model (BOTRReID \cite{luo2019bag}) to recognize the identity through body's features. 

The second stage, called the single-person behavior perception stage, processes the behavior of the identified characters separately. The gaze and head pose Estimation model (SYSUGaze\cite{li2022accurate}) are used to localize the direction of a character's head and gaze attention. The emotion recognition model \cite{yu2024joint} is used for three-category facial expression classification (neutral, happy, sad) and continuous valence and arousal emotion regression. The body key points model (HRNet\cite{sun2019deep}) is used to recognize hand-raising movements. The Yolov5\footnote{\url{https://github.com/ultralytics/yolov5}} model is used to detect the hand region and train a standard ResNet-50 model \cite{he2016deep} to recognize the 4-class gesture. The Automatic speech recognition model (Kaldi\cite{povey2011kaldi}) is used to recognize participants' speaking contents. 

We define the audio and video transcription task as a formulation of multiple perception model inference. This is denoted as:
\begin{equation}
\label{eq1}
[B]_{\mathrm{i}}=\mathrm{f}_{\text{image}}^{j}\left(I_{i}\right)+\mathrm{f}_{\text{audio}}^{\mathrm{k}}\left(S_{\mathrm{i}}\right)
\end{equation}

where $[B]_{\mathrm{i}}$ denotes the behavioral log of the $i_{\mathrm{th}}$ frame, $\mathrm{f}_{\text{image}}^{j}$ denotes the $j_{\mathrm{th}}$ image model, $\mathrm{f}_{\text{audio}}^{\mathrm{k}}$ denotes the $k_{\mathrm{th}}$ audio model, where $\left(I_{i}\right)$ and $\left(S_{i}\right)$ denote the $i_{\mathrm{th}}$ frame of image and audio signals, respectively.

To ensure a fair comparison with our previous work \cite{cheng2023computer}, we used the same but older perceptual model. More advanced ones can replace these ones to further the accuracy.

\begin{figure}[!h]
\centering
\centerline{\includegraphics[width=8.5cm]{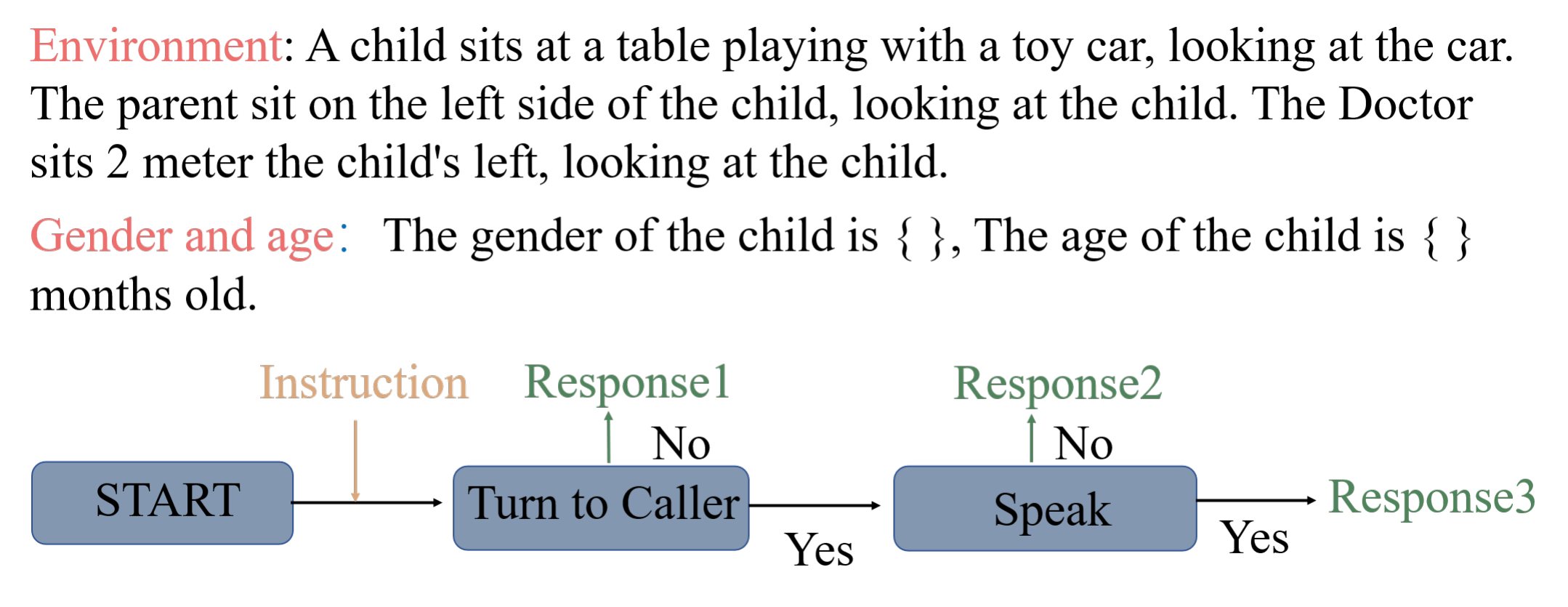}}
\caption{The textualization process of the Response to Name paradigm.}
\label{fig3}
\end{figure}

\subsection{Script Transcription Module}
Behavioral logs are generated by merging outputs from various models, resulting in data of different formats, such as body coordinates, eye vectors, facial expression types, gesture types, and speech content, etc.. To transform these behavioral logs into text that LLMs can understand, we developed a script transcription module to standardize and unify these diverse outputs into a coherent script.

\subsubsection{Response Parser}
Although the behavioral logs contain information about the actions of each individual in the video, they cannot be directly input into LLMs for analysis. There are two main reasons for this: 1. The perceptual model outputs either categories (e.g., facial expression categories) or values (e.g., gaze angles), and abstract inputs prevent LLMs from understanding the underlining behaviors. 2. Behavioral logs are recorded frame-by-frame, and lengthy inputs can lead to LLMs forgetting crucial information or struggling to detect ASD symptom representations. 

To address these issues, we need to utilize basic response events to simplify the behavioral log. In this paper, each video recording includes six paradigms, namely Response to Name (RN) \cite{frohna2007failure}, Social Smile (SS) \cite{qiu2020application}, Indicating Gesture (IG) \cite{zwaigenbaum2013early}, Responding to Joint Attention (RJA) \cite{talbott2020brief}, Initiating Joint Attention (IJA) \cite{mundy2003early}, and Separation Anxiety (SA) \cite{cheng2023computer}, these paradigms correspond to those presented in Table~\ref{tab:paradigm} and Table~\ref{tab:description}. As shown in Table~\ref{tab:paradigm}, We summarize the response events in the above paradigm that are most important for ASD observation. Generally, the observation of each paradigm is based on the following events. We define $E$ as the paradigm's set of all doctor-patient interactions. $E_{1}$ represents the event in which the child looks at the target object, called Look at Object. $E_{2}$ represents the event in which the child points at the target object, called Point to Object. $E_{3}$ denotes the event where the child smiles, called Smile. $E_{4}$ denotes the event when the child or doctor speaks, called Speak. $E_{5}$ represents the event in which parents or doctors exit the testing studio, called Leave. The following formulation defines the response parsing process:

\begin{equation}
\label{eq2}
R=E([B]) \text { w.r.t. } E=\left[E_{i}\right]_{i=1,2, \ldots 5}
\end{equation}
where $R$ records the timestamp of event $E$ occurring in $[B]$.

\begin{table*}[htb]
\caption{Instruction descriptions and response descriptions in the response textualization module}
\setlength{\tabcolsep}{1mm}
\label{tab:description}
\centering
\resizebox{\textwidth}{!}{
\begin{tabular}{lllll}
\toprule
\multicolumn{1}{c}{\textbf{Paradigm}} & \multicolumn{2}{l}{\textbf{ \hspace{6em} Instruction}} & \multicolumn{2}{l}{\hspace{8em} \textbf{Response}} \\ \cline{2-5} 
\multicolumn{1}{c}{}  & \multicolumn{1}{c}{\textbf{Session}} & \multicolumn{1}{c}{\textbf{description}} & \multicolumn{1}{c}{\textbf{name}} & \multicolumn{1}{c}{\textbf{description}} \\ \midrule
\multirow{3}{*}{RN} & P1 & The parent called out the child's name. & response 1 & The child turns toward the doctor and look  with saying hello. \\
                    & P2 & The doctor called out the child's name. & response 2 & The child turns toward the doctor and look . \\
                    &        &                                         & response 3 & The child continued to play with the toy. \\ \midrule
\multirow{7}{*}{SS} & P3 & The doctor greets the child with a passional smile and say hello. & response 1 & The child made eye contact with the doctor. \\
                    & P4 & The doctor praises the child with a warm smile. & response 2 & The child look at the doctor and smile . \\
                    & P5 & \makecell[l]{The doctor plays a tickle game with smile. She slowly reaches\\out and gently touches the child.} & response 3 &  The child smile  but did not look at the doctor. \\
                    & P6 & \makecell[l]{With a warm smile, parents entertain their children in whatever\\ way they normally do in their daily lives.} & response 4 & The child bent his head and went on playing with the toy. \\
                    &         &  & response 5 & The child made eye contact with the doctor without smile. \\ \midrule
\multirow{7}{*}{IG} & P7 & The doctor call the child's name and say "Look at that flower". & response 1 & The child looked up  in the direction of the picture. \\
                    & P8 & The doctor call the child's name and say "Look at that tree". & response 2 & The child keeps his head down and continues to play with his toy. \\
                    & P9 & The doctor call the child's name and say "Look at that balloon" & response 3 & The child precisely  points out the location of the picture. \\
                    & p10 & The doctor call the child's name and say "Look at that sofa" & response 4 & The child roughly points out the location of the picture. \\ 
                    &     &  & response 5 & The child turns around  and makes eye contact with the doctor. \\
                    &     &  & response 6 & The child keeps looking at the picture. \\
                    &     &  & response 7 & Then the kid continue to play with his toy. \\ \midrule
\multirow{5}{*}{RJA} & P11 & \makecell[l]{The doctor raises his hand and points to the picture of a clock \\ and says, "Look, there is a clock. what time it is.} & response 1 & \makecell[l]{The child turns his head backand\\ then looks to the position of the clock.} \\
                    &     &   & response 2 & The child seek the clock while not finding the correct direction. \\
                    &     &   & response 3 & The child looked up at the doctor's hand . \\
                    &     &   & response 4 & The child keeps his head down and continues to play with his toy. \\ \midrule
\multirow{9}{*}{IJA} & P12 & \makecell[l]{The wall to the left of the child suddenly displays a yellow\\ bird flapping its wings while a stereo plays the sound of bird.} & response 1 & \makecell[l]{The child is attracted to the animation playingand looks at the\\ bird on the left wall.}\\
                    & P13 & \makecell[l]{The wall to the right of the child suddenly displays a moving\\ riding car while the stereo plays the sound of the car moving.} & response 2 & \makecell[l]{The child turns around and makes eye contact with the doctor\\ to share his findings.} \\
                    & P14 & \makecell[l]{A cow wiggling its ears is suddenly displayed on the wall\\ behind the child's right side while the sound is played.} & response 3 & \makecell[l]{The child turns around and\\ makes eye contact with the doctor to share his findings.}\\
                    &  & & response 4 & The child keeps staring at the animation playing on the wall. \\
                    &  & & response 5 & The child raises his hand and points  to the bird on the wall. \\
                    &  & & response 6 & The child lower his head again and continued to play with the toy. \\ \midrule
\multirow{7}{*}{SA} & P15 & \makecell[l]{The parent gets up from their seat, walks past the child,\\ and finally leaves the room.} & response 1 & \makecell[l]{The child realizes that the parent has left and gets up and chases\\ him toward the door.} \\
                    & P16 & \makecell[l]{The parent call the child\'s name outside the door and say, \\ “Hi, mom is leaving. You have to play alone.} & response 2 & \makecell[l]{The child turns to the direction of the parent but remains seating\\ at the table.} \\
                    &     &   & response 3 & The child keeps his head down and continues to play with his toy. \\
                    &     &   & response 4 & The parents, the doctor and the child have left the room. \\
                    &     &   & response 5 & The child lower his head again and continued to play with the toy. \\ \midrule
\end{tabular}
}
\end{table*}
\subsubsection{Response Textualization Module}
After capturing a response event, we must describe its occurrence and behavior in the text format. The response textualization module can convert the predefined events in the paradigm video into textual descriptions. As shown in Fig.~\ref{fig3}, this illustrates the textualization process for the Response to Name paradigm. In the paradigm, the child participant is first guided to play with toys on the desk. Once their attention is engaged, the assessor suddenly calls the child's name from behind. The child can exhibit one of three responses: 1) no response, 2) turning to face the caller, and 3) turning to face the caller and responding verbally. The whole paradigm process is shown as a flowchart, and we select descriptions from the RN paradigm in Table~\ref{tab:description} for textualized combinations depending on the instructions and responses. In addition, we found that adding background and gender descriptions before the instruction description can improve diagnostic accuracy. Based on the paradigm's flowchart in our previous work \cite{cheng2023computer}, Table~\ref{tab:description} shows all paradigms' instruction and response descriptions. 

\subsubsection{Emotion Textualization Module}
The DSM-V \cite{regier2013dsm} emphasizes that "ASD children lack social or emotional reciprocity." This deficit manifests in their difficulty understanding the emotions of others, being unresponsive, or showing indifference to emotional signals. This could be visually reflected in facial expressions' valence or arousal values. However, the paradigm design in our dataset only considers category emotion classes (e.g., smile or neutral) in the Social Smile (SS) paradigm, as shown in Table~\ref{tab:paradigm}. To enable LLMs to better understand the deficits in emotional reciprocity, we aim to reflect continuous emotional dynamics in the script. Therefore, we designed an emotion textualization module to capture the emotional dynamic points and transform video segments near these points into textual emotion descriptions.

\begin{figure*}[ht]
    \centering
    \includegraphics[width=\textwidth,scale=1.00]{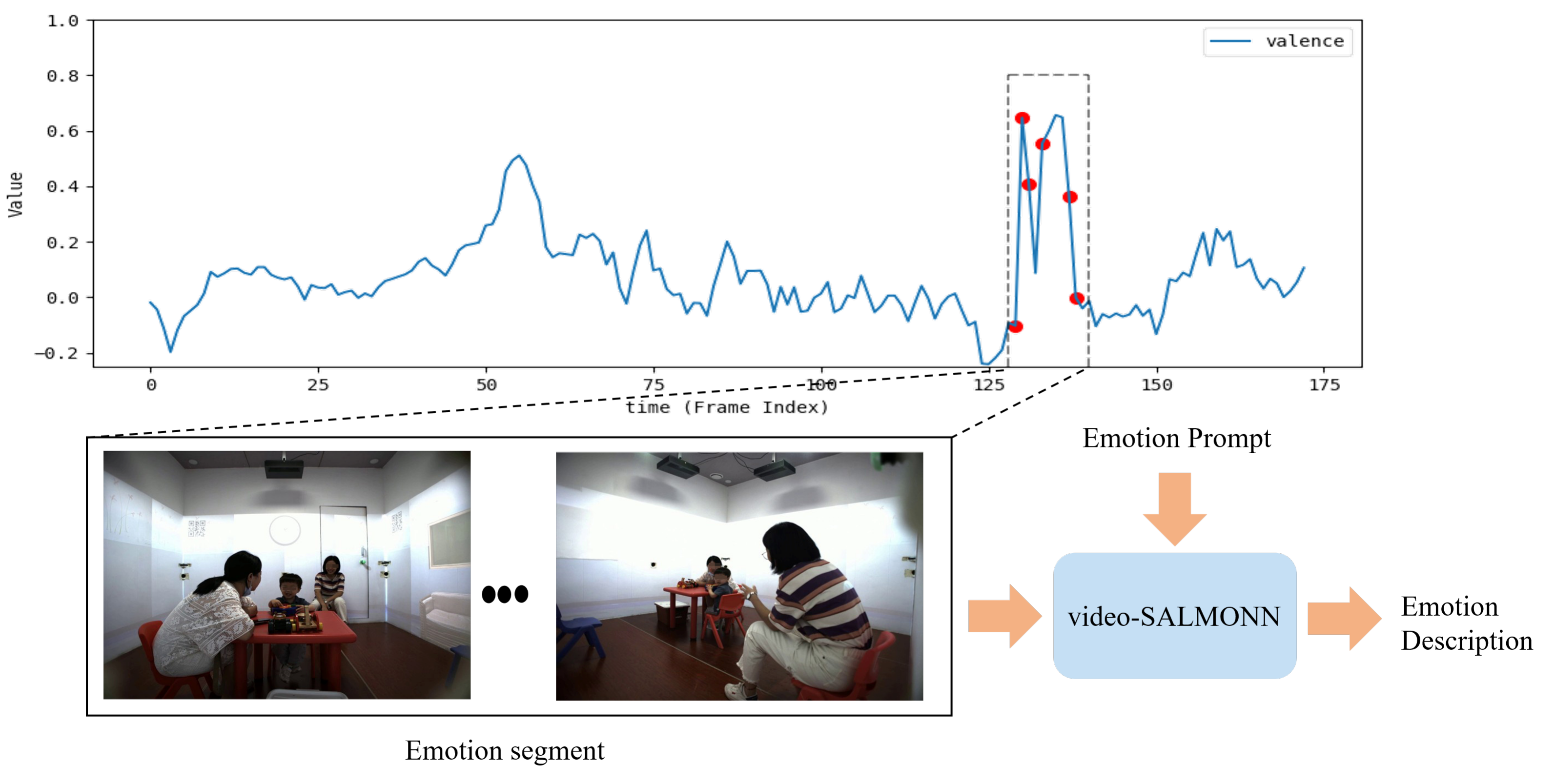}
    \caption{The emotion textualization process of the Response to Name paradigm. The blue line indicates the valence value. The red dotted line indicates the moment when the doctor call child's name. The red dots indicate points of emotional dynamic, and the black dashed interval highlights the segment where the child responded. } 
    \label{fig4}
\end{figure*}

The first-order derivative of emotion represents the rate of emotional change and serves as a key feature for understanding emotional dynamics \cite{Steephen2013,Krone2018,Zhu2024}. To capture emotional dynamic points, we approximate the degree of emotion variation using the first-order derivative of the valence value. The detailed process of finding emotional dynamic points is described in Algorithm \ref{alg:findemo}. Let $\left[a_1, a_2, \ldots, a_n\right]$ represent the valence sequence of facial expression, where $a_n$ is the valence value of the $n$th frame. Similarly, let $\left[d_1, d_2, \ldots, d_n\right]$ denote the sequence of first-order derivatives of the valence sequence, where $d_n$ represents the first-order derivative of the valence value at the $n$ th frame. We define $P$ as the emotional dynamic point and identify its location based on the emotional dynamic threshold $\alpha$. Specifically, we identify a $P_n$ when $d_n>0.2$ or $d_n<-0.2$ and define the 1-second video segment before and after this point as a emotional dynamic segment. Notably, a paradigm video may contain multiple emotional dynamic segments. If these segments overlap, they are merged into a single continuous mood segment. Finally, we get the

\begin{figure}[h!]
\centering
\centerline{\includegraphics[width=8.5cm]{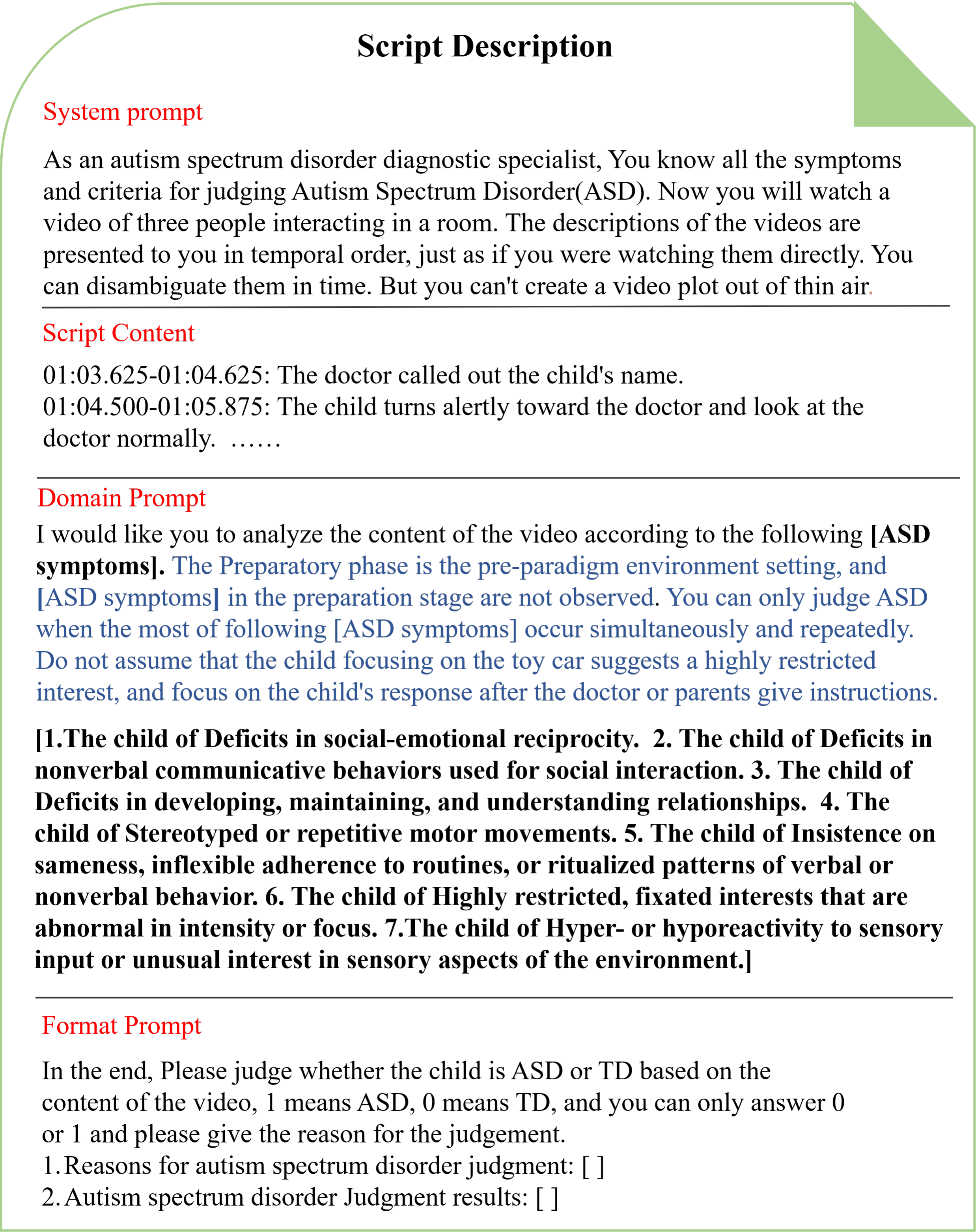}}
\caption{The details of the script description. The blue part is human experience. The bolded portion is domain knowledge}
\label{fig5}
\end{figure}

\noindent updated emotional dynamic points $P$ and their time intervals $[t_{n_{-} s t a r t}, t_{n_{-} e n d}]$. 

In transforming emotional fragments into emotional descriptions, it is challenging to manually summarize emotional dynamics into a limited set of discrete emotional events. In recent years, the general video understanding LLMs \cite{shu2024,qwen2.5vl,sun2024video} has demonstrated a remarkable ability to generate video descriptions, enabling the transformation of video content into emotionally relevant textual descriptions. Given that our data includes both video and audio modalities, we employ the audio-visual LLM model (video-SALMONN \cite{sun2024video}), to analyze visual frame sequences and audio events, with a primary focus on capturing emotionally relevant content. For segments, we generate textual descriptions of audio-visual content by prompting it with emotion-related queries. Figure~\ref{fig9} illustrates the emotion prompt and the emotion question. Fig~\ref{fig4} shows the complete process of the emotion textualization process. The valence line in the graph represents the emotional intensity of the child. This TD child responded by looking back at the doctor after hearing the command to call his name. The red dots on the reaction times indicate moments of large dynamics in valence. Clearly, the video shows the child's transition from a calm to a positive emotion.

\begin{figure*}[htb]
    \centering
    \includegraphics[width=\textwidth,scale=1.00]{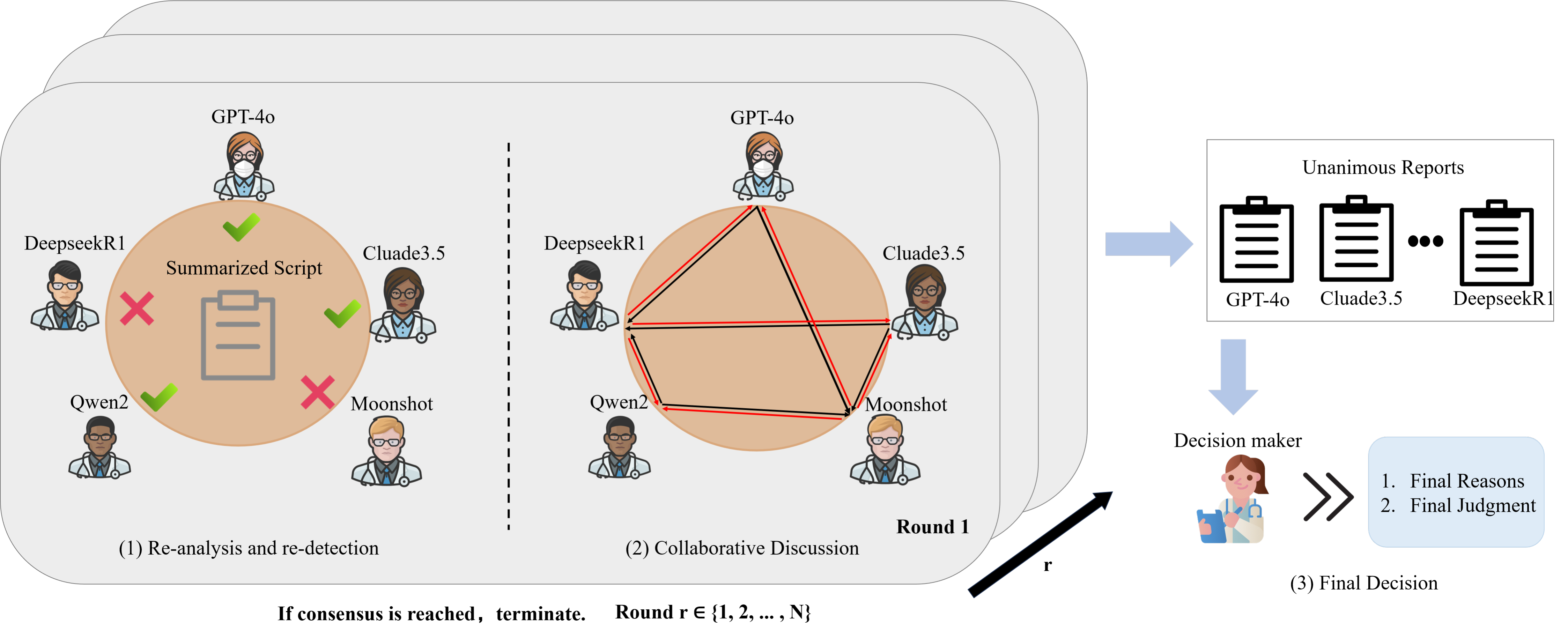}
    \caption{This is the framework for the joint detection of multiple LLMs. The process is divided into three stages: (1) Re-analysis and re-detection, (2) Collaborative Discussion, (3) Final Decision.} 
    \label{agents}
\end{figure*}

\subsubsection{Domain Prompt Module}
Prompt engineering can enhance LLMs' understanding of script descriptions. Therefore, we design a domain prompt module for ASD detection, as detailed in Fig~\ref{fig5}. The domain prompt module consists of four components: 1) The system prompt emphasizes the LLMs' identity to ensure they understand scripts in the temporal order. 2) The script content is generated by the response parser, the response textualization module and the emotion textualization. It records key events of the audio-visual behavior data in the text format as $\text{[timestamp, behavioral description]}$. 3) The domain prompt incorporates domain knowledge \cite{liu2023assessing} and experience into the script descriptions. In this context, domain knowledge refers to the ASD diagnostic criteria, while experience represents the researcher's experiential knowledge in the clinical setup. 4) The format prompt constrains the output results.

\subsection{Large Language Models}
In this paper, we selected eight LLMs with strong performance and reputation: three closed-source models (GPT-4O \cite{gpt4o}, Claude 3.5-sonnet \cite{Anthropic}, Monnshot \cite{Kimi}) and five open-source models (LLAMA 3.1 \cite{dubey2024llama}, qwen2-72B-instruct \cite{yang2024qwen2}, DeepSeek-R1-Distill-Qwen-32B \cite{guo2025deepseek},  DeepSeek-R1-Distill-Llama-70B \cite{guo2025deepseek}, DeepSeek-R1-671B \cite{guo2025deepseek}. We input script descriptions and user questions into these LLMs, and the assisted ASD detection results and interpretations were extracted from the answers. For few-shot learning, we only evaluate an open-source model deployed in our server because uploading data to a closed-source API may lead to data leakage. All LLMs are configured for fair experimentation with a $max\_token$ limit of 1000 and a $temperature$ setting of 0.7.

\subsection{The joint detection of multiple LLMs}
To further improve detection accuracy and reduce the hallucination from individual LLM, we fuse the outputs of multiple LLMs. The most straightforward strategy is a majority voting method, which we refer to as SCBU-Vote. We selecte five LLMs with high variability (GPT-4O \cite{gpt4o}, Claude 3.5-sonnet \cite{Anthropic}, Monnshot \cite{Kimi}, qwen2-72B-instruct \cite{yang2024qwen2}, DeepSeek-R1-671B \cite{guo2025deepseek}), and their predictions were aggregated through a voting mechanism to determine the final result. However, a clear limitation of this approach is that voting only applies to the final detection outcome, without providing a unified or interpretable rationale for the final decision.

Inspired by multi-agent medical diagnostics \cite{Tang2024}, we propose a framework for the joint detection of multiple LLMs called SCBU-Agents. The framework is divided into three stages, as shown in Fig~\ref{agents}: (1) Re-analysis and re-detection. Based on the behavioral scripts, the detection results of individual LLMs in the previous round, and the content of the discussion in the previous round,  the behavioral scripts are reanalyzed, and the ASD detection results are re-evaluated.  (2) Collaborative Discussion. Discuss the behavioral scripts with other LLMs and try to convince other LLMs who have different opinions.  If a consensus is reached or the maximum number of discussion rounds is exceeded, the process proceeds to Stage 3. Otherwise, it returns to Stage 1 for reanalysis. (3) Final Decision. We use GPT-4o to play the role of a decision maker and try to summarize the  LLMs' detection results of last round and get the final results.

\section{Experimental results}
\subsection{Dataset}
We utilized the multimodal behavioral database in \cite{cheng2023computer} to evaluate our method. This database comprises RGB, RGB-D and audio data recorded in a real clinical environment. Specifically, the dataset included eight distinct views of RGB HD video with a resolution of $4096 \times 3000$ and four different views of depth video with a resolution of $1280 \times 720$. These synchronized cameras ensured that the subject's behavior response were largely unobstructed.

\begin{figure}[htb]
\centering
\centerline{\includegraphics[width=8.5cm]{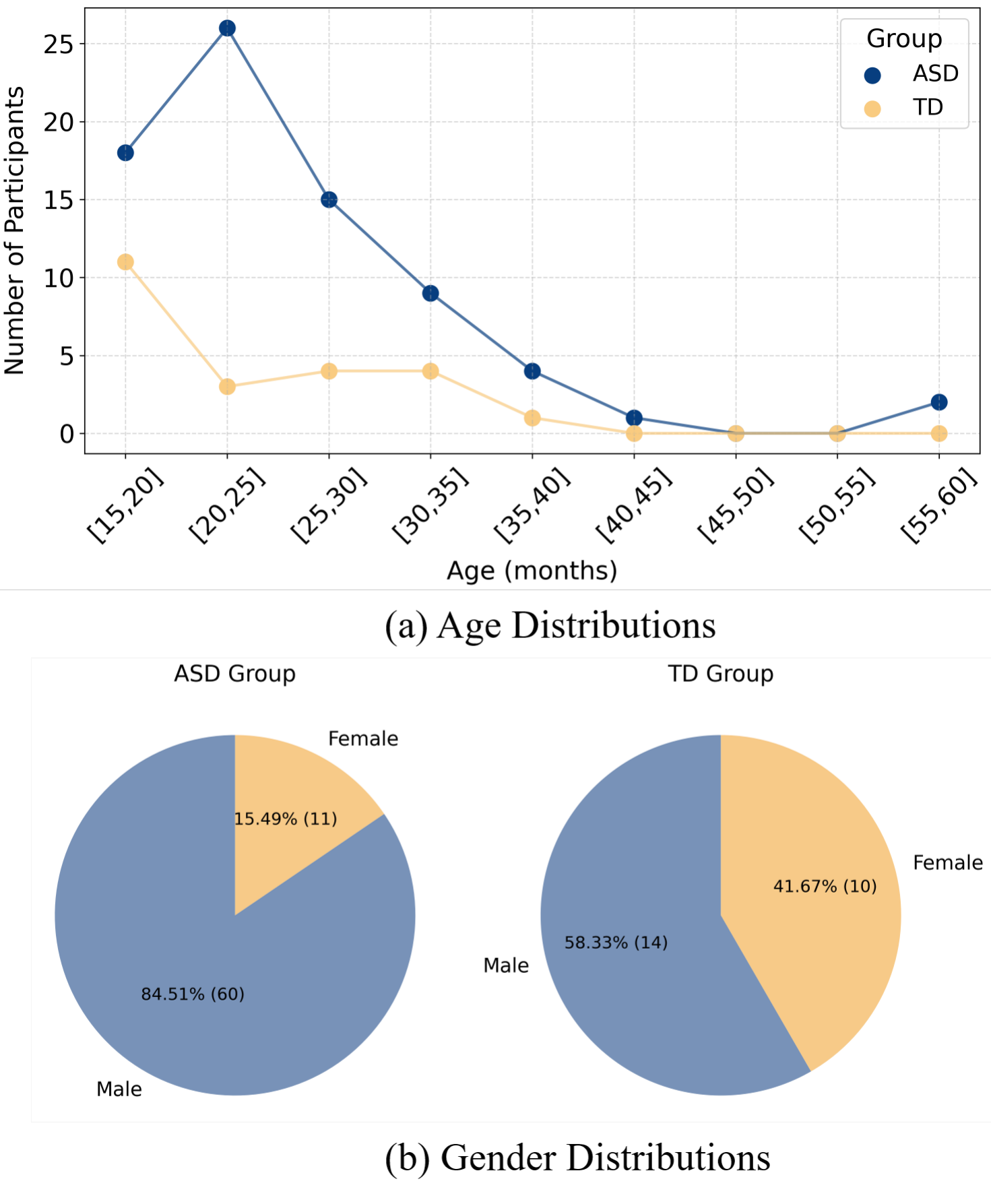}}
\caption{Statistics for ASD and TD groups in the clinical database. Scatterplot show age distribution of ASD and TD, pie chart show gender distribution of ASD and TD}
\label{fig6}
\end{figure}

The database included 95 participants, comprising 71 children diagnosed with ASD and 24 TD children. Fig.~\ref{fig6} illustrates the age and gender distributions of children with ASD and TD. The ages of ASD and TD participants in the dataset were clustered within the range of 15 to 30 months. The TD group exhibited a more balanced distribution between males and females, whereas the ASD group had a significantly higher proportion of males than females. These characteristics align with the clinical distribution of the data , More details about this dataset can be found in \cite{cheng2023computer}. For each assessment case, the physician will lead the child participant and parent through the six paradigms, which typically takes 20 to 30 minutes.

\begin{table}[htb]
\setlength{\tabcolsep}{1.1mm}
\caption{ASD detection results under the leave-one-out cross validation protocol. Supv denotes supervised learning.}
\label{tab:full_set}
\centering
\begin{tabular*}{\linewidth}{lcccccc}
\toprule
\textbf{Method} &\textbf{Supv}& \textbf{Emotion}&\textbf{ACC(\%)} & \textbf{F1(\%)}  & \textbf{SN(\%)} & \textbf{SP(\%)} \\ \midrule
Cheng et al. \cite{cheng2023computer}& $\surd$& - & 88.42 & 92.20  & 91.55  & 79.17     \\ \midrule 
SCBU-GPT4O& $\times$& w/o & 86.32 & 90.78  & 90.14  & 75.00  \\
          & $\times$& w/ & 88.42 & 92.62  & 97.18  & 62.50 \\  \midrule    
SCBU-Claude3.5& $\times$& w/o & 87.37 & 91.55  & 91.55 & 75.00     \\
             & $\times$& w/ & 87.37 & 91.18  & 87.32  & 85.75 \\  \midrule    
SCBU-Moonshot& $\times$& w/o  & 81.05 & 87.67  & 90.14 & 54.17    \\
            & $\times$& w/ & 84.21 & 88.55  & 81.69  & 91.67 \\  \midrule    
SCBU-Llama3.1&$\times$& w/o  & 82.11 & 88.89 &95.77& 41.67 \\ 
            & $\times$& w/ & 84.21 & 90.45  & \textbf{100.00}  & 37.50 \\  \midrule  
SCBU-Qwen2& $\times$& w/o  & 83.16 & 88.41  & 85.92  & 75.00     \\ 
            & $\times$& w & 87.37 & 91.18  & 87.52  & 87.50 \\  \midrule   
SCBU-DeepSeekR1& $\times$ &w/o  & 83.16 & 87.69  & 80.28  & 91.67       \\
-Distill-Qwen-32B & $\times$& w/ & 86.32 & 90.51 & 87.32  & 83.33 \\  \midrule   
SCBU-DeepSeekR1& $\times$ &w/o  & 85.26 & 90.67  & 98.77  & 54.17      \\
-Distill-Llama-70B & $\times$& w/ & 87.37 & 92.21 & 100.00  & 50.00 \\  \midrule
SCBU-DeepSeekR1& $\times$ &w/o  &  86.32 & 91.03  & 92.96 & 66.67     \\
-671B & $\times$& w/ & 90.53& 94.04 & 100.00 &  62.50 \\  \midrule
SCBU-Vote& $\times$ &w/o  & 86.32 & 90.28  & 91.55 & 70.83 \\ 
            & $\times$& w/ & \textbf{92.63}  & 94.44 &  95.77 & \textbf{88.33}  \\ \midrule
SCBU-Agents& $\times$ &w/o  & 87.37 & 91.78  & 94.37 & 66.67 \\ 
            & $\times$& w/ & \textbf{92.63}  & \textbf{95.24} &  98.59 & 75.00  \\ \bottomrule  
\end{tabular*}
\end{table}

\subsection{Overall Performance}
We employ four widely used metrics to evaluate methods in ASD detection: Accuracy (ACC), F1-score (F1), Sensitivity (SN), and Specificity (SP). Accuracy represents the percentage of correct ASD predictions. The F1-score considers both recall and precision, reflecting the overall performance of the methods. Sensitivity indicates the ability to identify ASD cases correctly, while specificity measures the ability to identify non-ASD cases correctly. Higher values for these metrics indicate better method performance.

Table~\ref{tab:full_set} compares the results of different LLMs and the impact of incorporating emotion descriptions. In addition, our proposed zero-shot method SCBU is also compared with the supervised method proposed by Cheng et al. in \cite{cheng2023computer}. When using a single LLM without emotion descriptions, SCBU-Claude 3.5 (w/o emotion) achieves an F1-score of 91.55\%, remarkably close to the supervised learning baseline (92.20\%). Similarly, SCBU utilizing other LLMs also achieves comparable accuracy, demonstrating the feasibility of LLMs for ASD detection. When using a single LLM with emotional descriptions, SCBU-DeepSeekR1-4O (w/ emotion) achieves an F1-score of 94.04\%, surpassing the supervised learning baseline. This result indicates that incorporating the emotional dynamic descriptions enhances the distinction between ASD and TD. Adding appropriate emotional descriptions in behavior scripts can further enhance LLMs' understanding of ASD. 

Finally, we fused the detection results of multiple LLMs using two strategies: SCBU-Vote and SCBU-Agents. The results of these fusion methods are presented in Table 1. Both approaches outperform all single LLM method and significantly surpass the supervised method. Notably, SCBU-Agents (w/ emotion) achieved the highest F1-score of 95.24\%. Further details of the SCBU-Agents’ discussion process and detection rationale are found at \url{https://github.com/lwx0724/Script-Centric-Behavior-Understanding}.

\begin{figure*}[htb]
    \centering
    \includegraphics[width=\textwidth,scale=1.00]{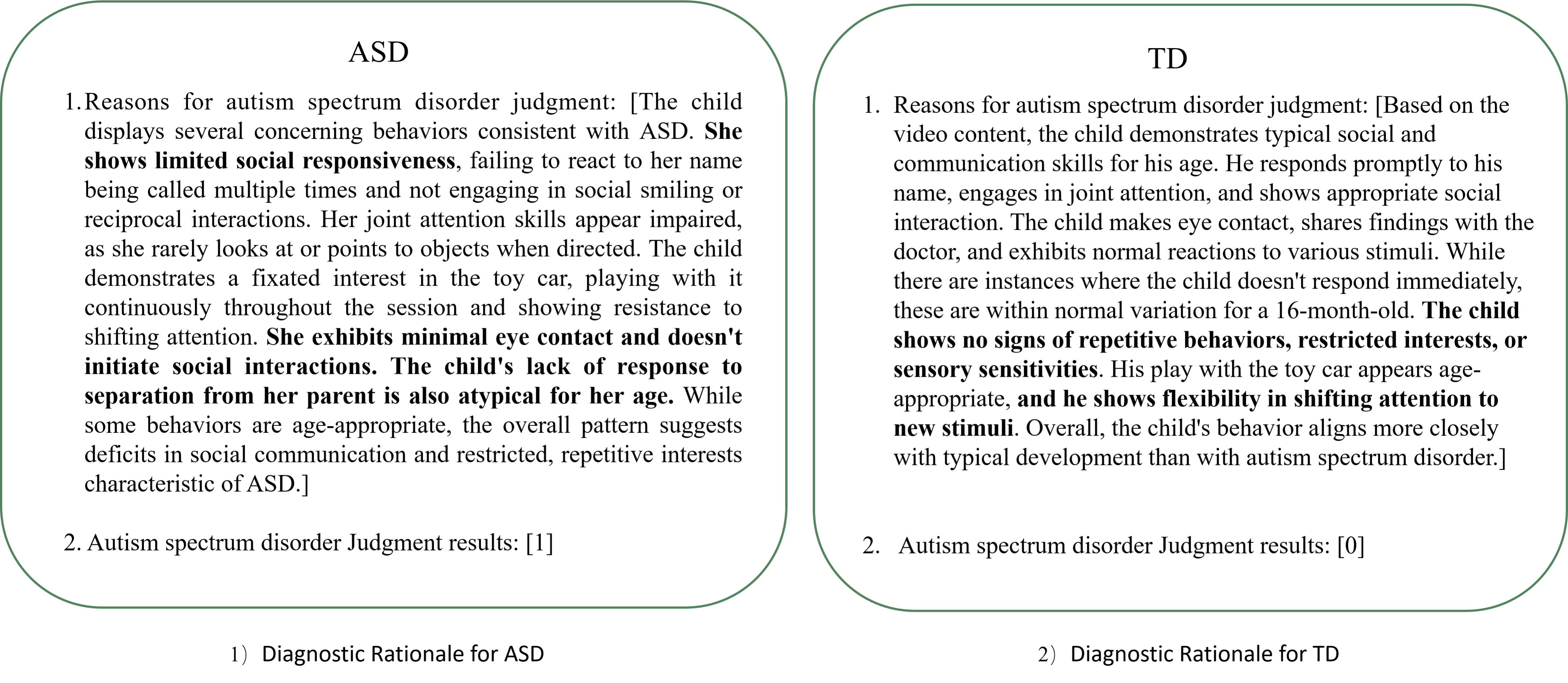}
    \caption{Examples of LLM detection for children with ASD and TD} 
    \label{fig7}
\end{figure*}

\begin{table}[htb]
\setlength{\tabcolsep}{1.5mm}
\caption{Comparison in valence, arousal, response Latency, response duration and dynamic frequency between the TD group and the ASD group}
\vspace{0.5em}
\label{tab:valence}
\centering
\begin{threeparttable}
\begin{tabular*}{\linewidth}{lcccc}
\toprule
\textbf{Dependent variable} &\textbf{TD Group}&\textbf{ASD Group} & \textbf{T Value}  & \textbf{P Value} \\ \midrule
Response Latency & & & &  \\
\hspace{1em} look & 4.9932 & 5.6257 & -2.5202 & 0.0154  \\
\hspace{1em} point &  8.5822 & 10.2493 & -3.0459 &  0.0042 \\
\hspace{1em} chase & 5.4036 & 10.0498 & -2.1848 & 0.0321  \\
Response Duration & & & & \\
\hspace{1em} look &  1.6503 & 1.3006 & 2.2138 & 0.0314  \\
\hspace{1em} point & 0.4355 & 0.2025 & -2.1848 & 0.0506 \\
\hspace{1em} chase & 10.3385 & 16.2931 & -5.6688& $<0.001$ \\
Valence Maximum& 0.3294 & 0.2840 & 5.3101 & $<0.001$     \\
Valence Minimum& -0.2905 & -0.2540 & -4.5054  & $<0.001$     \\
Arousal Maximum& 0.1977 & 0.1888 & 1.8850 & 0.0583    \\
Arousal Minimum& -0.1715 & -0.1648 &-1.5625 & 0.1184   \\
dynamic Frequency & 9.1150 & 6.8513 &  3.5492& $<0.001$ \\ 
dynamic Latency & 5.7320 & 6.7456 &  -2.1495 & 0.0319 \\ \bottomrule
\end{tabular*}
    \begin{tablenotes}    
    \footnotesize              
    \item[1] The Response Latency, the Response Duaration and the dynamic Latency are measured in seconds.         
    \item[2] The values of valence and arousal range from –1 to 1.  
    \item[3] The dynamic Frequency is measured in times.
  \end{tablenotes}           
\end{threeparttable}
\end{table}

In addition to achieving high diagnostic accuracy, Fig.~\ref{fig7} presents two examples demonstrating the interpretability of LLMs in autism detection. As shown in Fig.~\ref{fig7} 1), The LLMs' response include the reasons for the judgments and the judgment results. The reasons provided by the LLM for its judgments include competencies in social skills, responses to name-calling, smile interactions, stereotypical behaviors, and age factors. The bolded section on the left highlights the child's deficits in social interactions, which align with the DSM-V criteria for ASD diagnosis. As shown in Fig.~\ref{fig7} 2), This explains the LLM's detection in TD children. The LLM also considers ASD characteristics such as social and communication skills, attention, eye contact, desire to share, and age. In summary, the LLM's interpretation of the results aligns with human judgment expectations and can serve as an alternative physician reference. The results show that our method can explain the causes and enhance the credibility of assisted detection.

\subsection{Behavioral Distinction}
To verify whether behavioral descriptions can effectively distinguish ASD, Table \ref{tab:valence} demonstrates that all behaviors exhibited significant differences in response latency and response duration ($p < 0.05$). The mean delay in TD children looking at or pointing to the target object after the doctor's command was shorter than that in ASD children, while the mean duration was more prolonged. Furthermore, the mean latency in chasing ability was also shorter in TD children than in those with ASD. These findings align with ASD characteristics, such as reduced response flexibility and lower concentration in social interactions.

\subsection{Emotional Dynamic Distinction}
\begin{figure}[h!]
    \centering
    \centerline{\includegraphics[width=7.5cm]{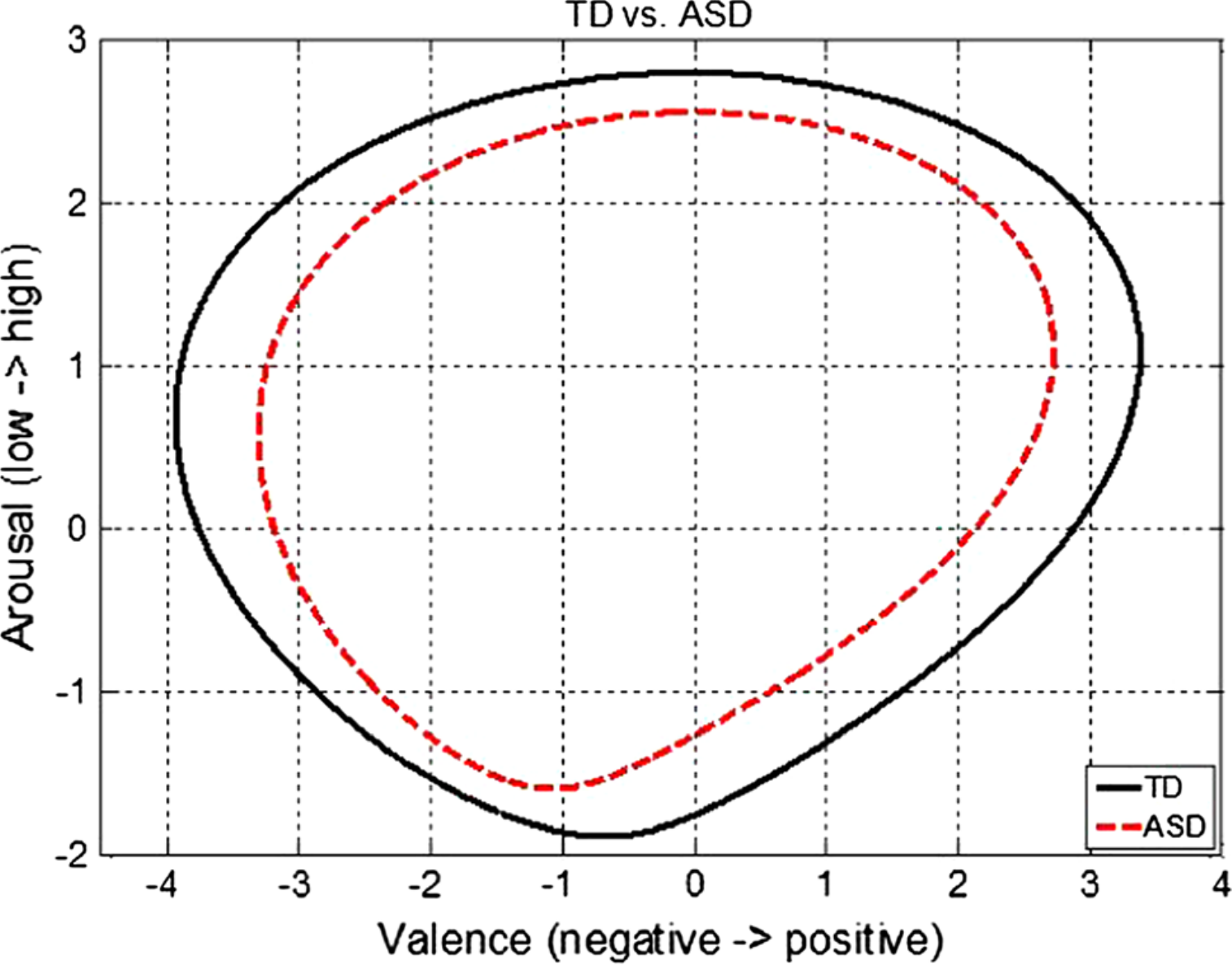}}
    \caption{Comparison in the ranges of valence and arousal between the ASD group and the TD group (duplicated from Fig. 3 in \cite{Tseng2013})} 
    \label{fig10}
\end{figure}

Valence and arousal are the two core dimensions of emotion, forming the fundamental framework of emotional representation. Valence determines the overall direction of emotion (positive or negative), while arousal describes its intensity\cite{Kuppens2013}. Tseng et al. \cite{Tseng2013} demonstrated that the range of emotional ratings in the ASD group was consistently limited. As shown in Fig.~\ref{fig10}, both the valence and arousal ranges of the ASD group were significantly smaller than those of the TD group. Therefore, we utilized the differences in valence and arousal values between the ASD and TD groups to identify the emotional dynamic points. Furthermore, a t-test was conducted on our data to determine which measure showed greater statistical significance. Table \ref{tab:valence} presents the significance analysis of valence and arousal. There was a less significant difference in the Arousal minimum between the TD and ASD groups ($p=0.1184$). Similarly, for the arousal maximum, the difference was also not high statistically significant ($p=0.0583$). In contrast, the TD group had a wider range of valence than the ASD group, and both the valence maximum and minimum differed significantly ($P < 0.001$). Hence, this paper utilizes valence values to describe the emotional dynamic points in children with ASD.

Why do the emotional dynamic points contribute to distinguish TD from ASD? Differences in emotional dynamics are demonstrated in Table \ref{tab:valence}. Children with ASD exhibit a lower frequency of emotional dynamics compared to TD children, with a statistically significant difference ($p < 0.001$). Additionally, emotional dynamics in children with ASD occurred later than in TD children, with a significant difference in latency ($p < 0.05$).

\begin{table}[ht]
\caption{ASD detection results for zero-shot evaluations with script descriptions generated by video-SALMONN.}
\vspace{0.5em}
\label{tab:General}
\centering
\begin{tabular*}{\linewidth}{lllll}
\toprule
\textbf{Method} &\textbf{ACC(\%)} & \textbf{F1(\%)}  & \textbf{SN(\%)} & \textbf{SP(\%)} \\ \midrule
GPT4O & 74.74 & 85.54  & 100.00  & 0.00  \\ 
Claude3.5& 75.79 & 85.89 & 98.59  & 8.33  \\
Moonshot & 74.74 & 85.54 & 100.00 & 0.00 \\
Llama3.1 & 74.74 & 85.54  & 100.00  & 0.00 \\  
Qwen2 & 74.74 & 85.37& 98.59& 00.04\\ 
DeepSeekR1-32B & 72.63& 83.33 & 91.55 & 16.67 \\
DeepSeekR1-70B & 73.68 & 84.85 & 98.59 & 0.00 \\
DeepSeekR1-671B & 74.74& 85.54 & 100.00 & 0.00\\ \bottomrule
\end{tabular*}
\end{table}

\begin{table}[h!]
\setlength{\tabcolsep}{1.0mm}
\caption{ASD detection results for few-shot evaluations with script descriptions generated by video-SALMONN}
\vspace{0.5em}
\label{tab:fewshot_SALMONN}
\centering
\begin{tabular*}{\linewidth}{lcccccc}
\toprule
\textbf{Method} & \textbf{Emotion} &\textbf{Few-Shot} &\textbf{ACC(\%)} & \textbf{F1(\%)}  & \textbf{SN(\%)} & \textbf{SP(\%)} \\ \midrule
\multirow{6}{*}{Qwen2} & \multirow{6}{*}{w/o} & 0-shot & 72.00  & 83.72 & 98.18 & 0.00     \\
&  & 4-shot & 73.33  & 84.62  & 100.00 & 0.00 \\
&  & 8-shot & 76.00  & 85.94  & 100.00 & 10.00 \\ 
&  & 12-shot & 74.67  & 85.27 & 100.00 & 5.00 \\ 
&  & 16-shot & 76.00  & 85.94  & 100.00  & 10.00   \\
&  & 20-shot & 77.33  & 86.61  & 100.00  & 15.00\\ \bottomrule
\end{tabular*}
\end{table}

\begin{table}[ht]
\setlength{\tabcolsep}{0.6mm}
\caption{ASD detection results under the 20-training and 75-testing few-shot evaluation protocol.}
\vspace{0.5em}
\label{tab:fewshot}
\centering
\begin{tabular*}{\linewidth}{lcccccc}
\toprule
\textbf{Method} & \textbf{Emotion} &\textbf{Few-Shot} &\textbf{ACC(\%)} & \textbf{F1(\%)}  & \textbf{SN(\%)} & \textbf{SP(\%)} \\ \midrule
cheng et al. \cite{cheng2023computer}  & -& 20-shot & 80.00  & 86.79  & 83.64 & 75.00     \\ \midrule
\multirow{6}{*}{SCBU-Qwen2} & \multirow{6}{*}{w/o} & 0-shot & 81.33  & 87.04  & 85.45 & 70.00     \\
&  & 4-shot & 85.33  & 90.27  & 92.73 & 70.00 \\
&  & 8-shot & 88.00  & 91.59  & 89.09 & 85.00 \\ 
&  & 12-shot & 89.33  & 92.98 & \textbf{96.36} & 70.00 \\ 
&  & 16-shot & 89.33  & 90.91  & 90.91  & 85.00   \\
&  & 20-shot & \textbf{92.00}  & \textbf{94.34}  & 90.91  & \textbf{95.00} \\ \midrule
\multirow{4}{*}{SCBU-Qwen2} & \multirow{4}{*}{w/} & 0-shot & 85.33  & 87.27  & 87.27 & 80.00     \\
&  & 4-shot & 85.67 &  89.72 & 87.27  & 80.00  \\ 
&  & 8-shot & \textbf{89.33} & \textbf{92.59} & \textbf{90.91}  & \textbf{85.00} \\ \midrule
\multirow{6}{*}{\makecell{SCBU-DeepSeekR1\\-Distill-Llama-70B}} & \multirow{6}{*}{w/o} & 0-shot & 82.67  & 88.89 & 94.55 &  50.00 \\
&  & 4-shot & 84.00  & 89.66  & 94.55 & 55.00 \\
&  & 8-shot & 86.67  & 91.38  & 96.36 & 60.00 \\ 
&  & 12-shot & 89.33  & 93.10 & 98.18 & 65.00 \\ 
&  & 16-shot &  90.67 & 93.91  & 98.18  & \textbf{70.00}  \\
&  & 20-shot &  \textbf{90.67} & \textbf{94.02}  & \textbf{100.00}  & 65.00 \\ \midrule
\multirow{4}{*}{\makecell{SCBU-DeepSeekR1\\-Distill-Llama-70B}} & \multirow{4}{*}{w/} & 0-shot & 84.00  & 90.16 & \textbf{100.00} &  40.00      \\
&  & 4-shot &  85.33 & 90.43  & 94.55  & 60.00  \\ 
&  & 8-shot & \textbf{89.33} & \textbf{92.98}& 96.36  & \textbf{70.00} \\ \bottomrule
\end{tabular*}
\end{table}

\subsection{Ablation Study}

\textbf{General video understanding model} Since generic large video understanding models can generate video descriptions, could they also facilitate the generation of script descriptions? To investigate this, we design the following ablation experiments. First, we use the video-SALMONN \cite{sun2024video} model to generate descriptions for each paradigm video. These descriptions were then chronologically integrated into script descriptions. Finally, we maintained the same domain prompts and LLMs when evaluating performance to ensure experimental fairness. As shown in Table \ref{tab:General}, the scripts generated by the video understanding model exhibited substantial performance degradation across various tested LLMs. Notably, the specificity of all models approaches zero, indicating that they have lost the ability to classify TD cases accurately. As shown in Table \ref{tab:fewshot_SALMONN}, We conducted a few-shot test using script descriptions generated by video-SALMONN, and the results similarly demonstrated the difficulty LLMs face in directly learning and analyzing from scripts generated by video understanding models. These experiments demonstrate the inability of LLMs to analyze scripts generated directly by video understanding models and underscores the importance of the transcription module we designed. A possible explanation for this is that behavior script descriptions capture critical responses of children with ASD during the paradigm process. In contrast, descriptions generated by generic video understanding models tend to be overly general, failing to capture the detailed discriminative behavior response.

\textbf{Few-Shot}  Given the fairness and label leakage concerns associated with closed-source LLMs that upload data, we conducted few-shot experiments on our locally deployed open-source models Qwen2 and DeepseekR1. Qwen2 has a maximum input token limit of 128,000, which constrains the length of the input script. After testing, we set the maximum few-shot number of scripts without emotional descriptions to 20. Since adding emotional descriptions increases the length of a single script, the maximum few-shot number is set as 8. To ensure experimental fairness, we divided the 95 participants into two groups: A few-shot training set include 20 children, consisting of 10 ASD and 10 TD. The test set include 75 children, consisting of 61 ASD and 14 TD. As shown in Table \ref{tab:fewshot}, the evaluation metrics of SCBU-Qwen2 (w/o Emotion)  continue to improve as the number of learning examples increases, ultimately reaching peak performance at a few-shot number of 20. It surpasses the supervised learning baseline proposed by Cheng et al. \cite{cheng2023computer} in all metrics. All experiments demonstrate that LLMs exhibit strong domain learning capabilities when provided with a few examples. Additionally, we conduct few-shot tests using DeepseekR1-Distill-Llama-70B (w/o Emotion) and DeepseekR1-Distill-Llama-70B (w Emotion). The results exhibit a trend consistent with Qwen2, with the best performance observed in the cases of DeepseekR1-Distill-Llama-

\begin{figure}[htb]
\centering
\centerline{\includegraphics[width=8.5cm]{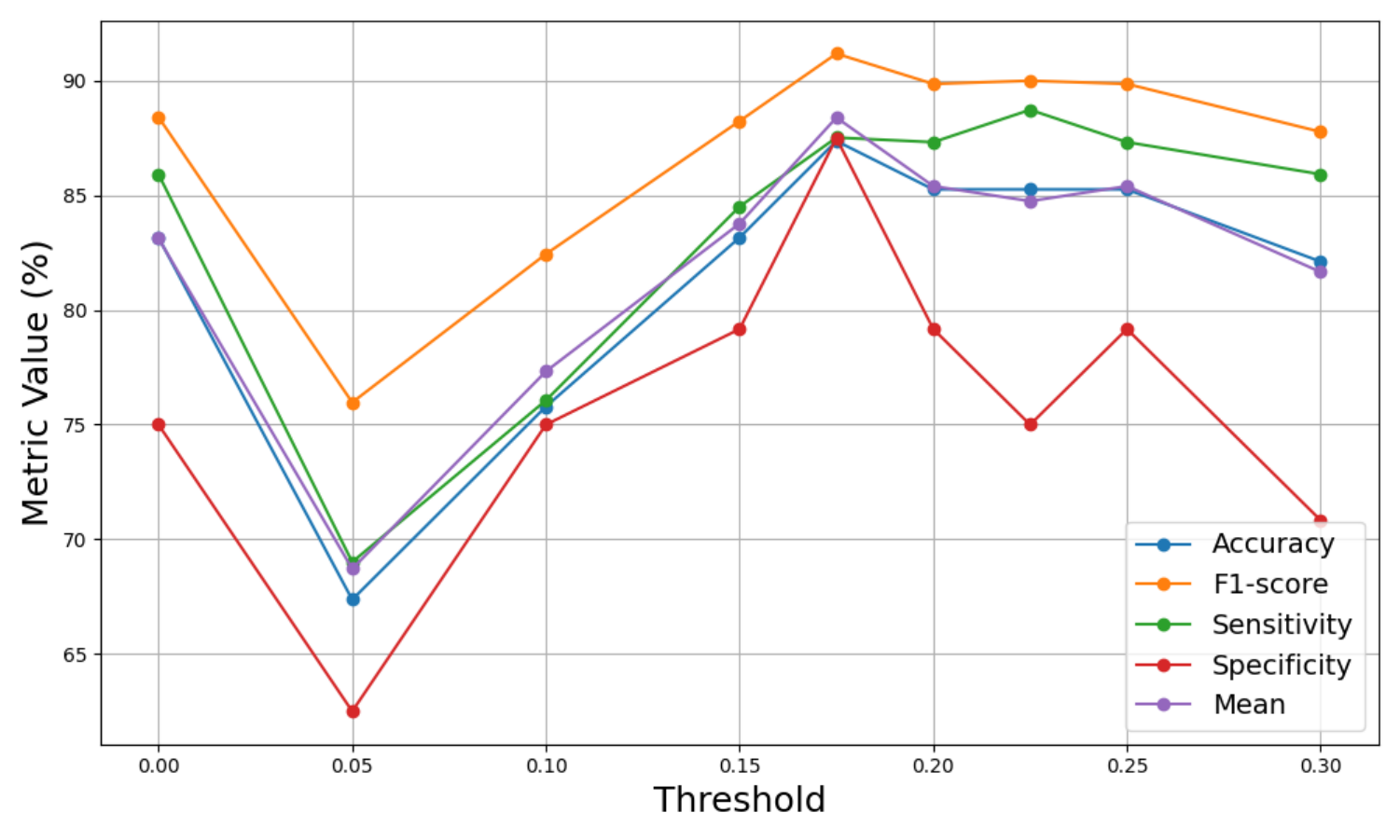}}
\caption{The impact of different emotional thresholds on different metric.}
\label{fig8}
\end{figure}

\begin{figure}[ht]
\centering
\centerline{\includegraphics[width=8.5cm]{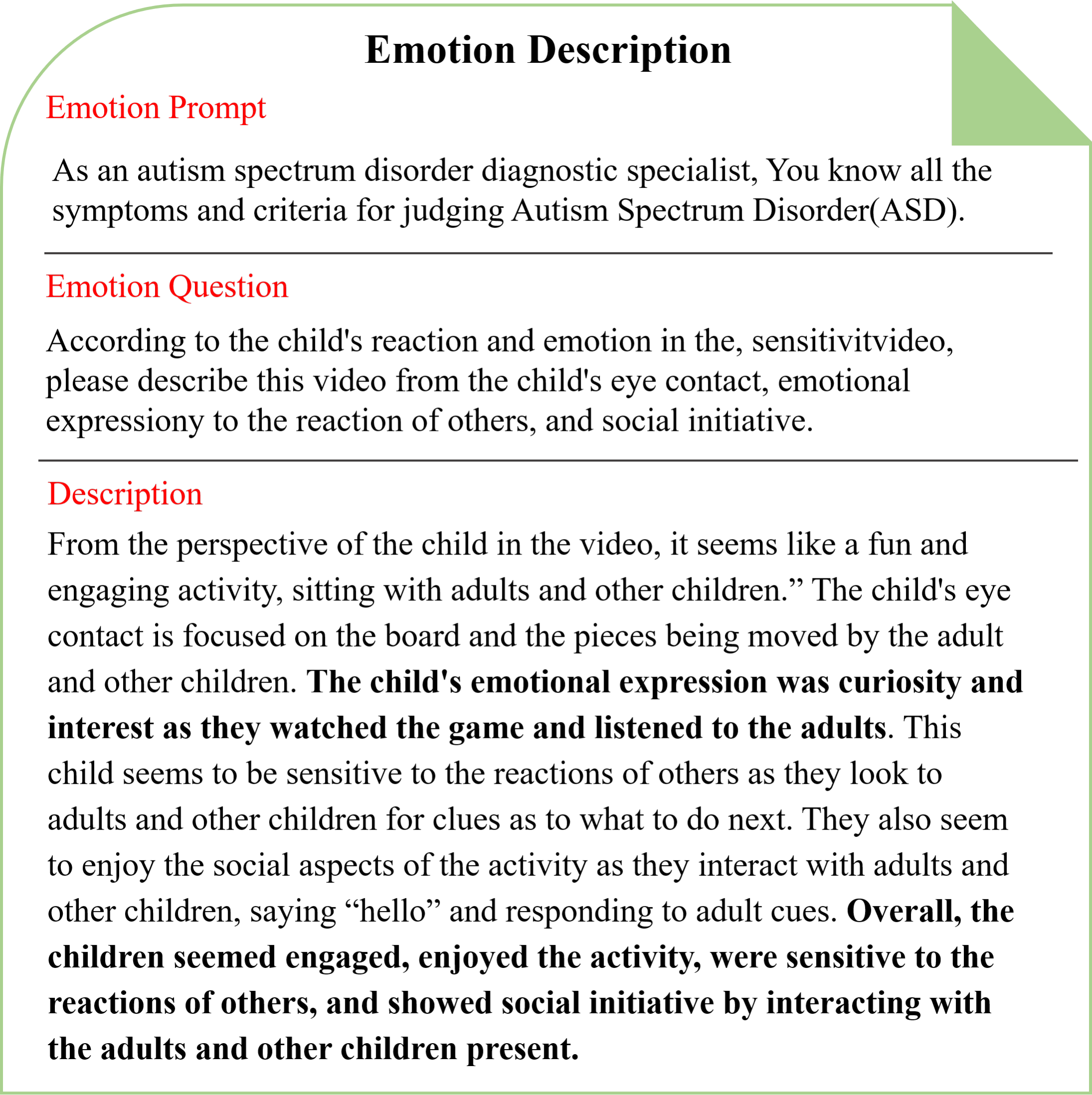}}
\caption{An example description of an emotional dynamic point.}
\label{fig9}
\end{figure}

\noindent 70B(w/o Emotion, 20-shot) and DeepseekR1-Distill-Llama-70B (w Emotion, 8-shot). These results validate the effectiveness of few-shot learning and underscore our method's potential in scenarios with limited training data. 

\textbf{Emotion thresholds}. The emotion threshold setting influences the number of emotional dynamic points. Specifically, a higher threshold results in fewer detected emotional dynamic points. If the number of emotional dynamics is too small, capturing variations in emotional valence becomes difficult. Conversely, an excessive number of dynamic points complicates the ability of LLMs to distinguish ASD from TD. Therefore, striking a balance between these factors requires careful selection of the emotion threshold. Fig~\ref{fig8} presents the evaluation metrics for different emotion thresholds.Thus, the emotion threshold for all experiments in this study is set to 0.175.

\begin{table}[ht]
\setlength{\tabcolsep}{1.2mm}
\caption{ASD detection results under the leave-one-out cross validation protocol with SCBU-GPT4O.}
\vspace{0.5em}
\label{tab:domain_prompt}
\centering
\begin{tabular*}{\linewidth}{cccccccc}
\toprule
\textbf{\makecell{No}}&\textbf{\makecell{Domain\\Knowledge}} & \textbf{\makecell{Human\\Experience}}& \textbf{\makecell{Emotion\\description}}& \textbf{ACC}& \textbf{F1} &\textbf{SN} &\textbf{SP} \\\midrule
 $\surd$ &          &         &  &74.74 & 84.00& 88.73&33.00\\
         &  $\surd$ &         &  &80.00 & 87.90 & 97.18&29.17 \\     
         &          & $\surd$ &  &77.89  &83.97 &77.46 & \textbf{79.17}    \\ 
         &          &         &$\surd$  &  77.89 & 86.65 & 94.37 & 29.17    \\ 
         & $\surd$  & $\surd$ &  & 84.21&89.36 &88.73 & 70.83 \\ 
         & $\surd$  &         &$\surd$  &83.16 & 89.47 & 95.77 & 45.83  \\ 
         &          & $\surd$ &$\surd$  & 85.26 & 90.00&88.73 & 75.00    \\ 
         & $\surd$  & $\surd$ & $\surd$ &\textbf{88.42} & \textbf{92.78}& \textbf{97.18} & 62.50   \\ \bottomrule
\end{tabular*}
\end{table}
\textbf{Emotion Description}. 
Fig~\ref{fig9} provides an example of an emotion description. This description corresponds to the emotional dynamics in Fig~\ref{fig8}. The emotion prompt makes LLM play the role of an ASD expert. The emotion question address several key aspects of ASD detection, such as eye contact, emotional expression, etc., ensuring a greater focus on emotionally relevant information related to ASD. The Description section in Fig~\ref{fig9} contains a textual description of the emotional dynamics observed in TD children. The bolded sections describe children's emotions, response sensitivity, and social initiative related to play, aligning with the actual events in the video. These suggest that the video-SALMONN model can describe emotion-related information.

\textbf{Domain Prompts Module}. To demonstrate the validity, we examined the impact of different prior information by ablation experiments: domain knowledge from DSM-V, human experience from researchers and emotion description from emotional dynamic point. As shown in Table~\ref{tab:domain_prompt}, the domain knowledge, the human experience, and emotion description can improve performance, respectively. Specifically, relying solely on domain knowledge can lead to high sensitivity but low specificity. However, incorporating human experience can effectively constrain the model, resulting in more balanced performance. Furthermore, adding emotional descriptions can comprehensively enhance the ASD detection capability of LLMs. These ablation experiments demonstrate the importance of designing tailored prompts for the ASD detection task. Our method achieved the best performance when all prior information are introduced simultaneously.

\subsection{Limitations and Discussions}
Our method converts audio-visual behavior into textual scripts to detect ASD. This conversion of clinical data into text helps prevent the leakage of patient privacy and facilitates data sharing among peers. Similarly, using text as a medium, our method adds descriptions of emotional dynamic points to the scripts, as if adding valid handcrafted features to existing data. The textual format also enables clinicians' experiences to be translated into behavioral features, making it easier to model practical behavioral markers. Additionally, the LLMs analyze the patient's behavioral script, enabling the model to provide diagnostic explanations, which assist doctors in analyzing the causes of ASD in children.

Although our method performs well in ASD detection, it still has limitations in practical scenarios. First, the detection requires both audio and video to be recorded in a controlled environment, as it is essential to ensure that the quality of the data is sufficient for stable behavior analysis. Second, the diagnostic video content is limited to structured paradigms, and robust modules for behavior transcription are still not mature in unstructured settings. Finally, while emotional descriptions enhance the effectiveness of LLMs in ASD detection, they depend on the model's ability to understand audio-visual data. The current general video understanding models still have limitations in psychological analysis. In future work, we aim to enable behavior perception and transcription in unstructured scenes or train a generalized video understanding model explicitly focused on behavior response, so LLMs can better perform behavior assessment tasks in the ASD domain.

\section{CONCLUSION}
 In this study, we propose a novel zero-shot and few-shot approach for detecting ASD using LLMs. We develop a script transcription Module to convert audio-visual content into text scripts. We design a domain prompts module to better leverage prior knowledge of ASD. Furthermore, we added an emotion textualization module to convert videos with intense emotional dynamics into textual descriptions to enhance the quality of behavior. Extensive experimental results demonstrate the effectiveness of our method, showing strong zero-shot and few-shot capabilities. Moreover, LLMs explain the reasoning behind ASD detection, which helps physicians analyze the detection process more effectively. Future research will focus on developing more straightforward and generalized behavioral perception models for video in unconstrained environments or general video understanding models in the ASD domain.

\bibliography{IEEEabrv,bibliography}


\begin{IEEEbiography}
[{\includegraphics[width=1in,height=1.25in,clip,keepaspectratio]{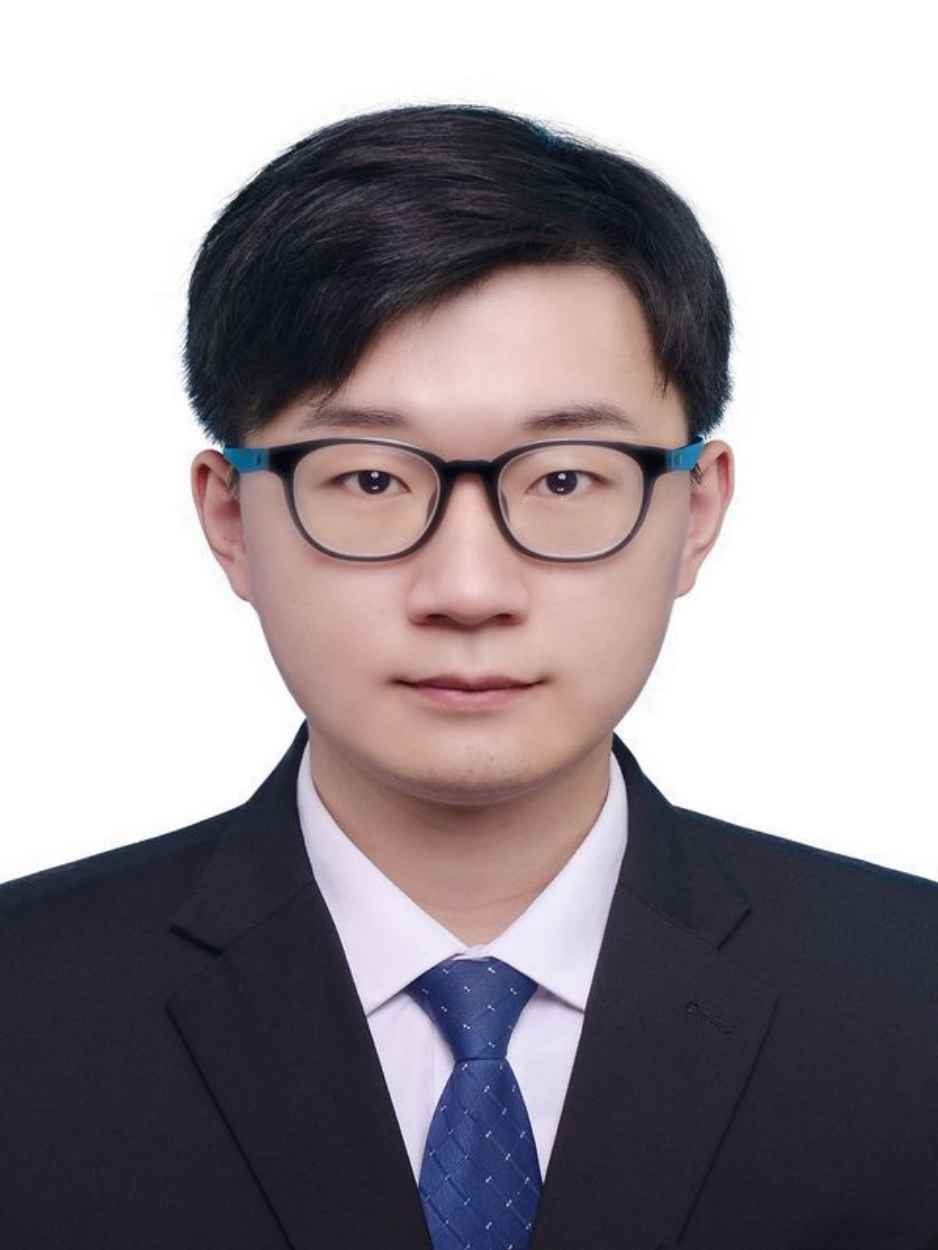}}]{Wenxing Liu}
received the master’s degree in computer science from the Chongqing University of Technology. He is currently working toward the PhD degree in computer science with Wuhan University. His research interests include ASD diagnosis, ASD assessment, and multimodal large language model.
\end{IEEEbiography}

\begin{IEEEbiography}
[{\includegraphics[width=1in,height=1.25in,clip,keepaspectratio]{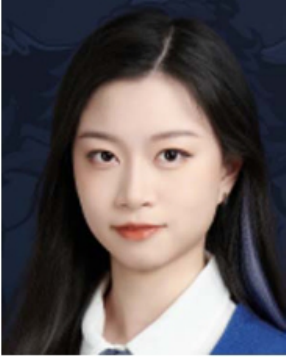}}]{Yueran Pan}
received the bachelor’s degree in statistics from Wuhan University, and the master’s degree in data science with distinction from the London School of Economics and Political Science. She is currently working toward the PhD degree in computer science with Wuhan University. Her research interests include applications of multimodal behavior analysis to help children with autism.
\end{IEEEbiography}

\begin{IEEEbiography}[{\includegraphics[width=1in,height=1.25in,clip,keepaspectratio]{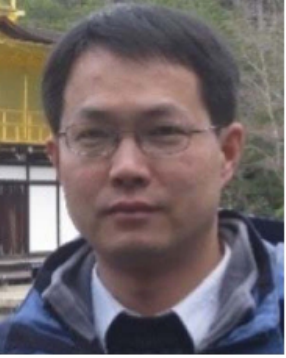}}]{Dong Zhang}
received the BSEE and MS degrees from Nanjing University, in 1999 and 2003, respectively, and the PhD degree from Sun Yat-sen University, in 2009. He is currently an associate professor with the School of Electronics and Information Technology, Sun Yat-sen University. His research interests include image processing, computer vision, affective computing, and information hiding.
\end{IEEEbiography}

\begin{IEEEbiography}[{\includegraphics[width=1in,height=1.25in,clip,keepaspectratio]{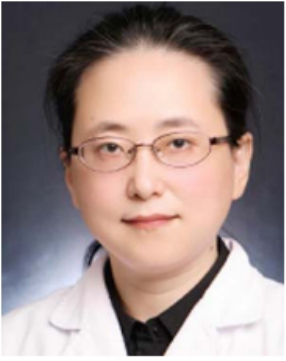}}]{Hongzhu Deng}
received the PhD degree in medicine from Sun Yat-sen University, in 2010. She is currently an associate professor with the Department of Pediatrics and the director with the Child Development and Behavior Center, Third Affiliated Hospital of Sun Yat-sen University, Guangzhou, China. Her research interests include the diagnosis and treatment of autism and other developmental disabilities.
\end{IEEEbiography}

\begin{IEEEbiography}[{\includegraphics[width=1in,height=1.25in,clip,keepaspectratio]{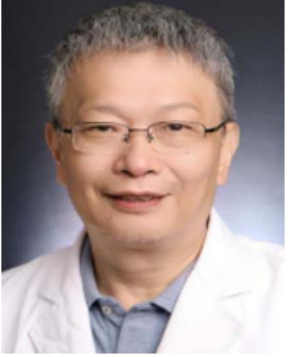}}]{Xiaobing Zou}
is the academic leader with the Child Development and Behavior Center, Third Affiliated Hospital of Sun Yat-sen University, Guangzhou, China. He has rich clinical experience in the field of developmental behavioral disorders for children. His research focuses on early diagnosis and behavioral intervention strategies for children with the spectrum and other developmental disabilities.
\end{IEEEbiography}

\vspace{-5 pt} 

\begin{IEEEbiography}[{\includegraphics[width=1in,height=1.25in,clip,keepaspectratio]{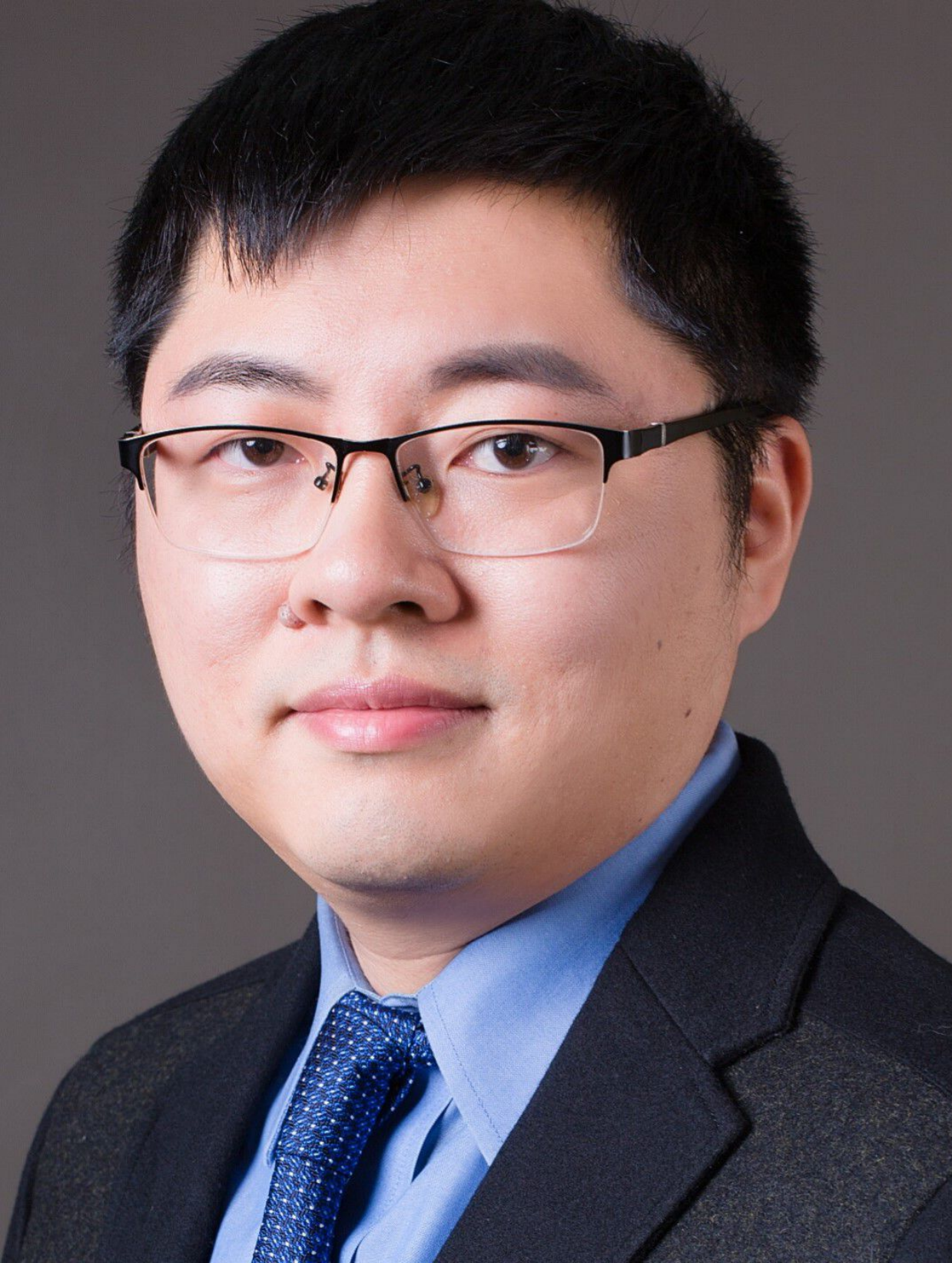}}]{Ming Li}(Senior Member, IEEE)
received his Ph.D. in Electrical Engineering from University of Southern California in 2013. He is currently a Professor of Electronical and Computer Engineering at Duke Kunshan University. He is also an Adjunct Professor at School of Computer Science in Wuhan University. His research interests are in the areas of audio, speech and language processing as well as multimodal behavior signal processing. He has published more than 200 papers and served as the member of IEEE speech and language technical committee, APSIPA speech and language processing technical committee, the editorial board member of the IEEE/ACM Transactions on Audio, Speech, and Language Processing and Computer Speech \& Language. He is an area chair at Interspeech 2016, 2018, 2020 and 2024, 2025 as well as the technical program co-chair of Odyssey 2022 and ASRU 2023. Works co-authored with his colleagues have won first prize awards at Interspeech Computational Paralinguistic Challenges 2011, 2012 and 2019, ASRU 2019 MGB-5 ADI Challenge, Interspeech 2020 and 2021 Fearless Steps Challenges, VoxSRC 2021, 2022 and 2023 Challenges, ICASSP 2022 M2MeT Challenge, IJCAI 2023 ADD challenge, ICME 2024 ChatCLR challenge and Interspeech 2024 AVSE challenge. He received the IBM faculty award in 2016, the ISCA Computer Speech and Language 5-years best journal paper award in 2018 and the youth achievement award of outstanding scientific research achievements of Chinese higher education in 2020.
\end{IEEEbiography}

\end{document}